\pdfoutput=1
\documentclass[10pt,twocolumn,letterpaper]{article}

\usepackage{wacv}
\usepackage{times}
\usepackage{epsfig}
\usepackage{graphicx}
\usepackage{amsmath}
\usepackage{amssymb}
\usepackage{booktabs}
\usepackage{multirow}
\usepackage{adjustbox}
\usepackage{pifont}
\usepackage{floatrow}
\usepackage[table]{xcolor}
\usepackage{paralist}
\usepackage{float}
\usepackage{comment}
\usepackage{authblk}
\floatstyle{plaintop}

\def\wacvPaperID{1425} 

\wacvalgorithmstrack   

\wacvfinalcopy 

\usepackage{ifthen}

\newcommand{\cmark}{\ding{51}}%
\newcommand{\xmark}{\ding{55}}%
\newcommand{\boldstart}[1]{\noindent\textbf{#1}}
\newcommand{\boldstartspace}[1]{\vspace{0.1in}\noindent\textbf{#1}}
\newcommand{\mat}[1]{\textbf{#1}}

\definecolor{third}{rgb}{1,1, 0.6}
\definecolor{second}{rgb}{1, 0.8, 0.6}
\definecolor{best}{rgb}{1, 0.6, 0.6}
\newcommand{\cb}{\cellcolor{best}}
\newcommand{\cs}{\cellcolor{second}}
\newcommand{\ct}{\cellcolor{third}}

\newcommand{\final}{1}  

\ifthenelse{\equal{\final}{1}}
{
\newcommand{\KL}[1]{}
\newcommand{\YC}[1]{}
\newcommand{\jason}[1]{}
}
{
\newcommand{\KL}[1]{{\color{red}[KL: #1]}}
\newcommand{\YC}[1]{{\color{blue}[YC: #1]}}
\newcommand{\jason}[1]{{\color{blue}[Jason: #1]}}
}

\ifwacvfinal
\usepackage[breaklinks=true,bookmarks=false,hidelinks]{hyperref}
\else
\usepackage[pagebackref=true,breaklinks=true,colorlinks,bookmarks=false]{hyperref}
\fi

\begin{document}

\title{Vision Transformer for NeRF-Based View 
Synthesis from a Single Input Image}

\makeatletter
\renewcommand*{\Authsep}{\hspace{0.10\textwidth}}
\renewcommand*{\Authand}{\hspace{0.10\textwidth}}
\renewcommand*{\Authands}{\hspace{0.10\textwidth}}
\renewcommand\AB@affilsepx{\hspace{0.10\textwidth} \protect\Affilfont}
\makeatother

\author[1]{Kai-En Lin\thanks{Work done while interning at Google.}}
\author[2]{Lin Yen-Chen}
\author[3]{Wei-Sheng Lai}
\author[4]{Tsung-Yi Lin\thanks{Work done while at Google.}}
\author[3]{Yi-Chang Shih}
\author[1]{Ravi Ramamoorthi}
\affil[1]{UC San Diego}
\affil[2]{MIT}
\affil[3]{Google}
\affil[4]{NVIDIA}



\maketitle
\thispagestyle{empty}

\begin{abstract}
Although neural radiance fields (NeRF) have shown impressive advances in novel view synthesis, most methods require multiple input images of the same scene with accurate camera poses.
In this work, we seek to substantially reduce the inputs to a single unposed image. 
Existing approaches using local image features to reconstruct a 3D object often render blurry predictions at viewpoints distant from the source view.
To address this, we propose to leverage both the global and local features to form an expressive 3D representation.
The global features are learned from a vision transformer, while the local features are extracted from a 2D convolutional network.
To synthesize a novel view, we train a multi-layer perceptron (MLP) network conditioned on the learned 3D representation to perform volume rendering.
This novel 3D representation allows the network to reconstruct unseen regions without enforcing constraints like symmetry or canonical coordinate systems.
Our method renders novel views from just a single input image, and generalizes across multiple object categories using a single model.
Quantitative and qualitative evaluations demonstrate that the proposed method achieves state-of-the-art performance and renders richer details than existing approaches.
\url{https://cseweb.ucsd.edu/\%7eviscomp/projects/VisionNeRF/}
\end{abstract}

\section{Introduction}

\begin{figure}[t]
\centering
\includegraphics[width=1.0\textwidth]{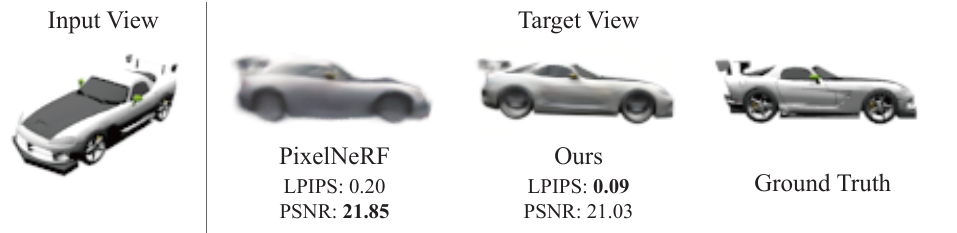}
\caption{\boldstart{Novel view synthesis in occluded regions.} 
    The visual quality of image-conditioned model (\eg, PixelNeRF~\cite{yu2020pixelnerf}) degrades significantly when pixels in the target view are invisible from the input.
    We propose to incorporate both global features from vision transformer (ViT) and local appearance features from convolutional networks to achieve significantly better rendering quality with more details in the occluded regions.
    Note that LPIPS~\cite{zhang2018perceptual} (lower is better) reflects the perceptual similarity better than PSNR.}
\label{fig:teaser}
\end{figure}

\begin{figure}[t]
\includegraphics[width=1.0\textwidth]{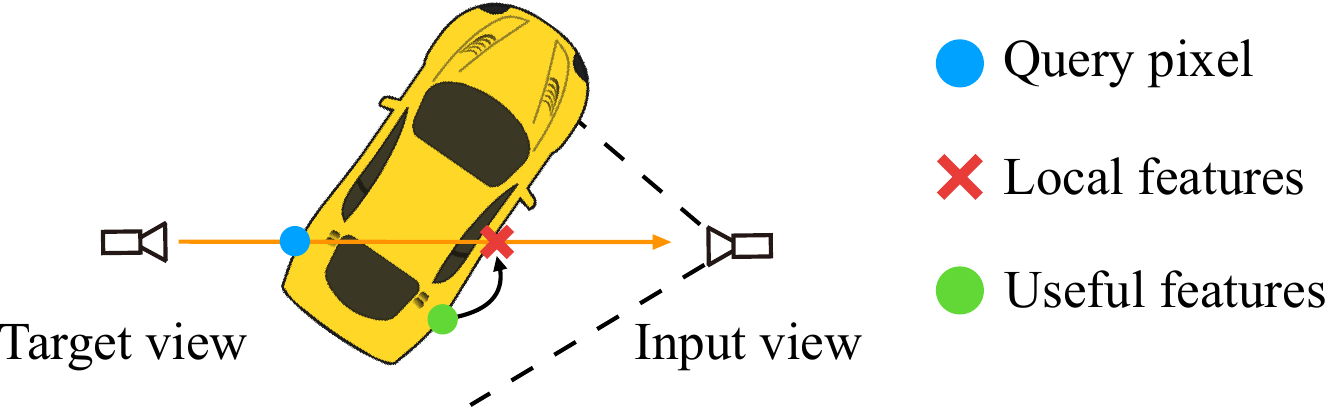}
  \caption{\boldstart{The challenge of image-conditioned models in the presence of self-occlusion.} 
  To render a car's occluded wheel (blue dot) in the target view, image-conditioned models, \eg, PixelNeRF~\cite{yu2020pixelnerf}, query features along the ray, which corresponds to the car's window in the input view (red cross).
  Our method uses self-attention to learn long-range dependencies, which is able to find the most related features in the source view (green dot) for rendering a clear target view.
  %
  }\label{fig:attention}
\end{figure}

We study the problem of novel view synthesis from a \textit{single unposed image}.
Recent works~\cite{reizenstein2021common,pifuSHNMKL19,yu2020pixelnerf} infer the 3D shape and appearance by projecting the input image features on the queried 3D point to predict the color and density.
These image-conditioned models work well for rendering target views close to the input view.
However, when target views move further, it causes significant occlusion from the input view, leading to dramatic degradation of the rendering quality, as shown in Fig.~\ref{fig:teaser}.
%
We hypothesize that self-occlusion causes the incorrectly-conditioned features.
As illustrated in Fig.~\ref{fig:attention}, when the query pixel in the target view (\eg, the car's wheel) is invisible from the input view, image-conditioned models incorrectly use the features from other surface (\eg, the car's window) for the target view.

To tackle this issue, we propose a novel approach that utilizes the recent advances in vision transformer (ViT)~\cite{dosovitskiy2020vit} and neural radiance fields (NeRF)~\cite{mildenhall2020nerf} to learn a better 3D representation.
We first lift the input 2D image into feature tokens and apply ViT to learn global information.
Subsequently, the feature tokens are unflattened and resampled into multi-level feature maps which allow the network to capture global information in a coarse-to-fine manner.
In addition, we adopt a 2D convolutional neural network (CNN) to extract local features that capture details and appearance from the input image. 
Finally, we render the novel viewpoints using the volumetric rendering technique~\cite{mildenhall2020nerf}.
Our method is able to render unseen regions with more accurate structure and finer details.

We train and evaluate our method on the ShapeNet dataset~\cite{shapenet2015} including 13 object categories.
Our method generalizes well across multiple categories, and works well on real-world  images. 
Quantitative and qualitative comparisons demonstrate that our method performs favorably against existing approaches, \eg, SRN~\cite{sitzmann2019srns}, PixelNeRF~\cite{yu2020pixelnerf}, FE-NVS~\cite{fast-and-explicit-neural-view-synthesis}, SRT~\cite{srt22}, and FWD~\cite{Cao2022FWD}, and generates more visually appealing results.
%
We summarize our contributions as follows:
\begin{compactitem}
    \item We introduce a NeRF-based rendering method that synthesize novel views from a single unposed image.
    \item We propose a novel 3D representation that integrates global and local information using vision transformer and 2D CNN. 
    \item We demonstrate state-of-the-art performance against existing approaches on category-specific and category-agnostic datasets as well as real input images.
\end{compactitem}

\section{Related work}

\begin{table*}[t]
  \centering
  \caption{\boldstart{Comparisons with recent novel-view synthesis methods.} 
  Our method takes as input a single image to perform novel view synthesis.
  Different from methods that assume an object-centered coordinate system, we infer the 3D representation in viewer-centered coordinate system and thus do not require the camera pose of the input.
  Additionally, our method is able to generalize to multiple categories using a single model.
  We extract local image features using 2D CNN and retrieve global information using a ViT encoder to synthesize faithful and appealing details on occluded regions (see Fig.~\ref{fig:teaser}).}
  \resizebox{\linewidth}{!}{
  \scriptsize
  \begin{tabular}{l c c c c c c c c c}
  \toprule
                    & NeRF & PIFu & PixelNeRF & CodeNeRF & NeRFormer & FE-NVS & SRT & FWD & \multirow{2}{*}{\textbf{Ours}} \\
                    & \cite{mildenhall2020nerf} & \cite{pifuSHNMKL19} & \cite{yu2020pixelnerf} & \cite{jang2021codenerf} & \cite{reizenstein2021common} & \cite{fast-and-explicit-neural-view-synthesis} & \cite{srt22} & \cite{Cao2022FWD} & \\
    \midrule
    Single-view input             & \xmark & \cmark & \cmark & \xmark & \xmark & \cmark & \cmark & \cmark & \cmark \\
    Viewer-centered coordinate    & \xmark & \cmark & \cmark & \xmark & \cmark & \cmark & \cmark & \cmark & \cmark \\
    Cross-category generalization & \xmark & \cmark & \cmark & \xmark & \cmark & \cmark & \cmark & \cmark & \cmark \\
    Image features                & \xmark & \cmark & \cmark & \xmark & \cmark & \cmark & \xmark & \cmark & \cmark \\
    Global features               & \xmark & \xmark & \xmark & \cmark & \xmark & \xmark & \cmark & \xmark & \cmark \\
    \bottomrule
  \end{tabular}}
  \label{tab:method_comp}
\end{table*}


\subsection{Novel View Synthesis}
Earlier works in view interpolation~\cite{10.1145/166117.166153} and light fields~\cite{gortler1996lumigraph,levoy1996light} establish the groundwork for image-based rendering.
Later works utilize proxy geometry~\cite{buehler2001unstructured,debevec1996modeling} and layered representations~\cite{shade1998layered,szeliski1999stereo} to better represent the 3D scene and synthesize novel views.
There has been a plethora of learning-based methods~\cite{flynn2019deepview,flynn2016deepstereo,kalantari2016learning,lin2021deep,lombardi2019neural,mildenhall2019llff,sitzmann2019deepvoxels,zhou2018stereo} and single-input view synthesis algorithms~\cite{niklaus20193d,pifuSHNMKL19,Shih3DP20,wu2020unsup3d,NIPS2019_8340,10.1145/3306346.3323007}.
These approaches exploit the differentiable rendering pipeline to generate photorealistic results.
Recently, neural radiance fields (NeRF)~\cite{mildenhall2020nerf} encodes the 3D scene in a compact continuous 5D function, allowing photorealistic reconstruction of the given scene.
Nonetheless, it requires tens or hundreds of input images and time-consuming optimization to train a single scene.
To address this problem, several methods~\cite{reizenstein2021common,grf2020,wang2021ibrnet,yu2020pixelnerf} utilize 2D image features to improve the generalization, or use pretrained networks with 1D latent code to represent the 3D shape, \eg CodeNeRF~\cite{jang2021codenerf}.
Guo et al.~\cite{fast-and-explicit-neural-view-synthesis} adopt a discrete 3D volume to represent the scene and achieve real-time rendering performance.
Instead of relying on pure 1D, 2D, or 3D representations, we propose to learn a novel 3D representation that utilizes global information and local image features.
Table~\ref{tab:method_comp} compares the proposed method to previous approaches.

\subsection{Transformer}
The transformer architecture~\cite{NIPS2017_3f5ee243} has brought significant advances in natural language processing (NLP).
While self-attention and its variant have achieved state-of-the-art performance in many NLP~\cite{brown2020language,devlin2018bert} and vision~\cite{dosovitskiy2020vit,Ranftl2021,strudel2021} tasks, directly applying self-attention to an image is prohibitively expensive, as it requires each pixel to be attended to every other pixel.
%
Several works~\cite{hu2019local,parmar2018image,ramachandran2019stand,zhao2020exploring} approximate self-attention by applying it to local patches of each query pixel.
Recently,the vision transformer (ViT)~\cite{dosovitskiy2020vit} and follow-up works~\cite{Ranftl2021,wang2021pyramid} demonstrated that applying a transformer to a sequence of patches (split from an image)  achieves competitive performance on discriminative tasks (\eg, image classification).
%
Wang et al.~\cite{wang2021multi} include transformers in both the encoder and decoder for 3D reconstruction from multi-views.
NeRF-ID~\cite{arandjelovic2021nerf} uses a transformer to sample 3D points along rays.
Other approaches~\cite{johari2021geonerf,reizenstein2021common,wang2021ibrnet} use transformers to aggregate source view features extracted by a CNN. 
Our work is different from these methods as we focus on learning global image information using ViT.
In our experiment, ViT encodes image features that achieves higher reconstruction quality on unseen regions than previous CNN-based approaches.
SRT~\cite{srt22} uses a fully transformer-based framework to encode and decode 3D information.
It learns the 3D scene information as a set of latent code, while our work adopts radiance field as the scene representation.
SRT uses a transformer to decode the set of latent code, whereas our method uses the per-pixel information from a set of feature maps, thus having an explicit mapping between the input image and the 3D point query.
Sec.~\ref{sec:category_agnostic} shows that our proposed method achieves favorable results over SRT in PSNR and SSIM metrics.


\section{Novel View Synthesis From a Single Image}

Our goal is to infer a 3D representation from a single input image for novel view synthesis.
We first discuss three different paradigms to learn such a 3D representation (Sec.~\ref{sec:representation}).
Then, we propose a hybrid representation to improve rendering quality on occluded regions, where we utilize a ViT to encode global information (Sec.~\ref{sec:vit}) and a 2D CNN to encode local appearance features (Sec.~\ref{sec:cnn}).
Finally, we learn a NeRF~\cite{mildenhall2020nerf} module that conditions the encoded features for novel view synthesis (Sec.~\ref{sec:volume_rendering}).

\begin{figure}[t]
\centering
\includegraphics[width=1.0\textwidth]{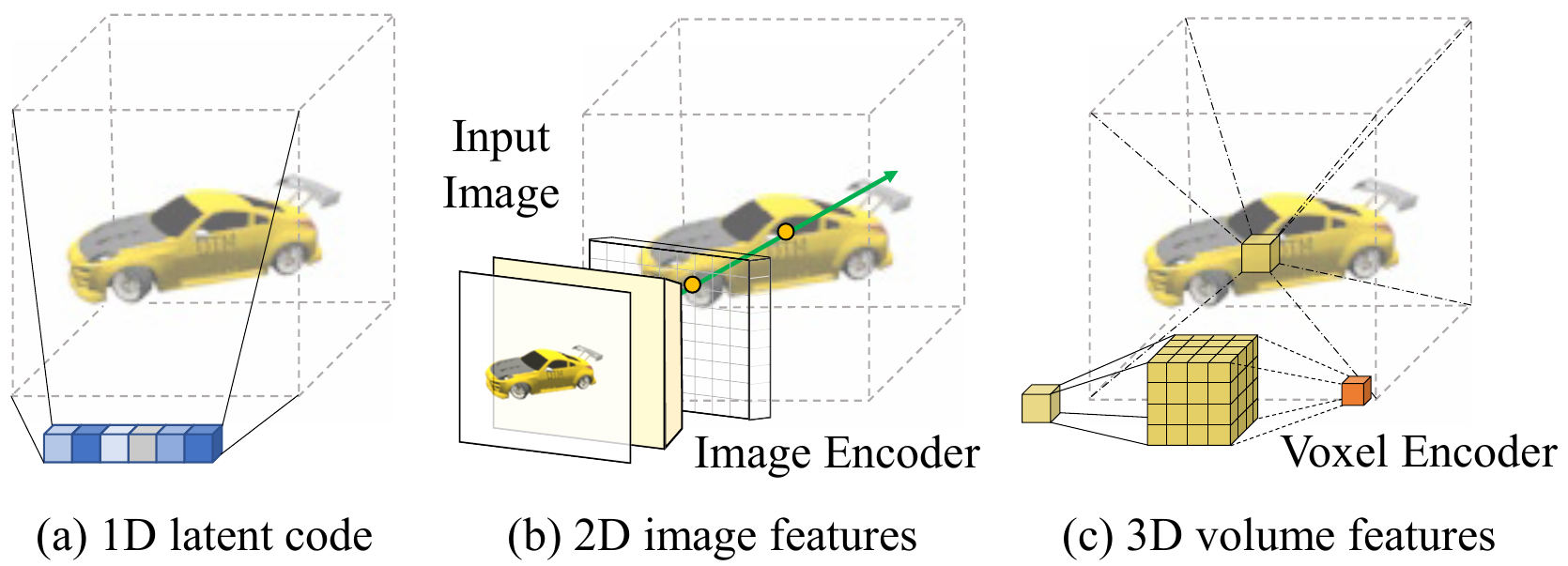}
\caption{\boldstart{Illustration of different representations for a 3D object.} 
(a) 1D latent code-based approaches~\cite{chen2018implicit_decoder,dupont2020equivariant,jang2021codenerf,Occupancy_Networks,DVR,Park_2019_CVPR} encode the 3D object in an 1D vector. 
(b) 2D image-based methods~\cite{pifuSHNMKL19,yu2020pixelnerf} are conditioned on the per-pixel image features to reconstruct any 3D point. 
(c) 3D voxel-based approaches~\cite{fast-and-explicit-neural-view-synthesis,lombardi2019neural} treat a 3D object as a collection of voxels and apply 3D convolutions to generate color and density vector RGB$\sigma$.}
\label{fig:representation}
\end{figure}

\begin{figure*}[t]
\centering
\includegraphics[width=1.0\textwidth]{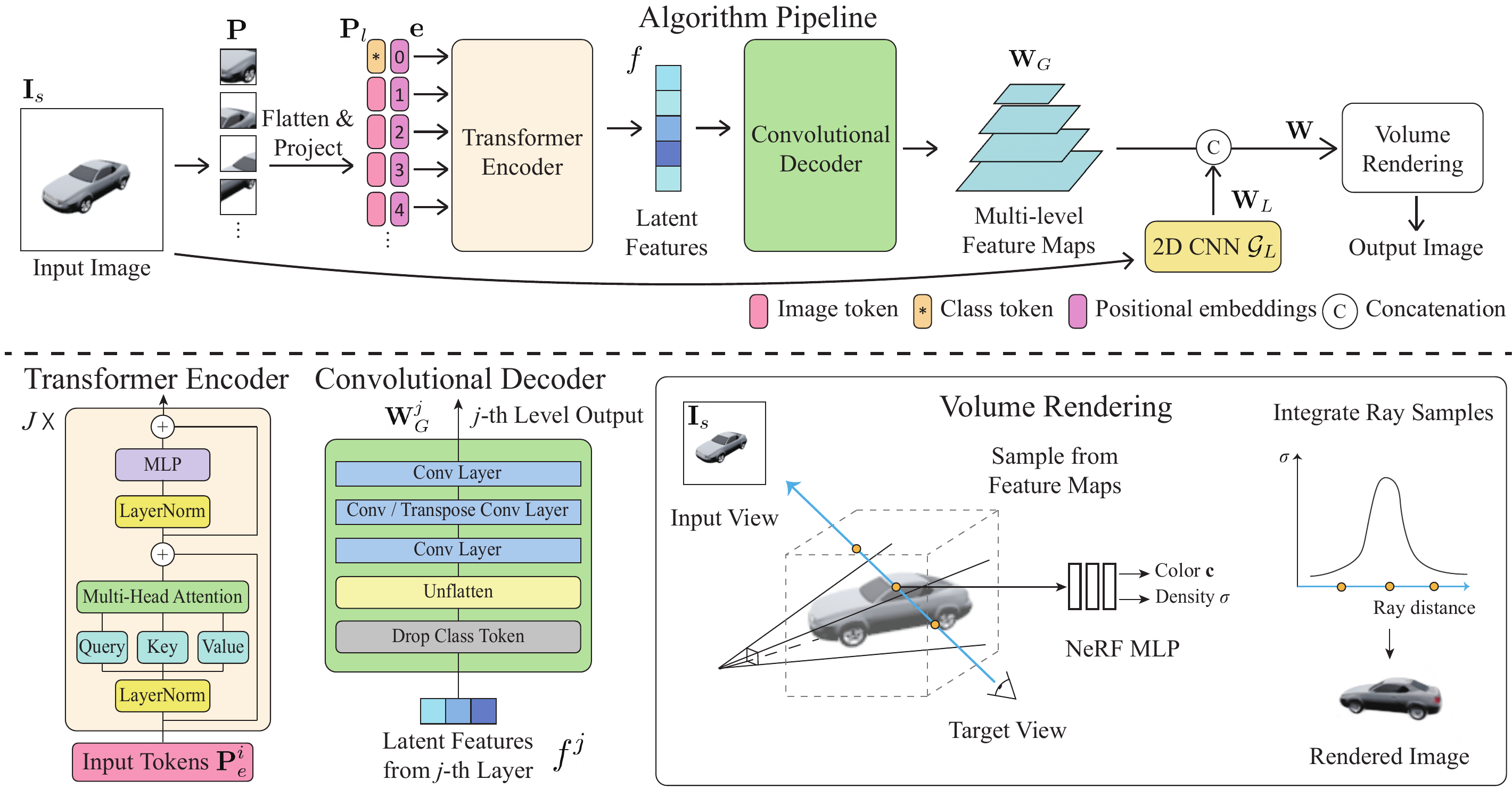}
\caption{\boldstart{Overview of our rendering pipeline.} 
We first divide an input image into $N = 8 \times 8$ patches $\mat{P}$. 
Each patch is flattened and linearly projected to an image token $\mat{P}_l$. 
The transformer encoder takes the image tokens and learnable positional embeddings $\mat{e}$ as input to extract global information as a set of latent features $f$ (Sec.~\ref{sec:vit}). 
Then, we decode the latent feature into multi-level feature maps $\mat{W}_G$ using a convolutional decoder.
In addition to global features, we use another 2D CNN $\mathcal{G}_L$ to obtain local image features (Sec.~\ref{sec:cnn}). 
Finally, we sample the features for volume rendering using the NeRF MLP (Sec.~\ref{sec:volume_rendering}).}
\label{fig:algorithm}
\end{figure*}

\subsection{Synthesizing Occluded Regions}\label{sec:representation}
In this section, we describe how previous works and our method reconstruct unseen regions illustrated in Fig.~\ref{fig:representation}.
Additionally, we analyze the strengths and weaknesses of each method, and propose a hybrid representation to address the critical issues in existing methods.
Given a single image $\mat{I}_s$ at camera $s$, our task is to synthesize novel view $\mat{I}_t$ at camera $t$.
If a 3D point $\mat{x}$ is visible in the source image, we can directly use the color $\mat{I}_s(\pi(\mat{x}))$, where $\pi$ denotes the projection to source view, to represent the point as seen by a novel viewpoint.
If $\mat{x}$ is occluded, we resort to information other than the color at the projection $\pi(\mat{x})$.
There are three possible solutions to gather such information.

\boldstartspace{1D latent code.}
Existing methods encode 3D and appearance prior through a 1D global latent vector $\mat{z}$~\cite{chen2018implicit_decoder,dupont2020equivariant,jang2021codenerf,Occupancy_Networks,DVR,Park_2019_CVPR,rematas2021sharf}, and decode the color $\mat{c}$ and density $\sigma$ through CNN as the following, shown in Fig.~\ref{fig:representation}(a)~\cite{jang2021codenerf}:
\begin{equation}
    (\sigma, \mat{c}) = \mathcal{F}_{\text{1D}}(\mat{z}; \mat{x}; \mat{d}).
\end{equation}
where $\mat{x}$ and $\mat{d}$ denotes the the spatially-varying sampling position and viewing direction.
Since different 3D points share the same latent code, the inductive bias is limited.

\boldstartspace{2D spatially-variant image feature.}
There are many interests around image-conditioned methods, such as PIFu~\cite{pifuSHNMKL19} and PixelNeRF~\cite{yu2020pixelnerf}, due to the flexibility and high-quality results around the input views.
These approaches are more computationally efficient as they operate in the 2D image space rather than 3D voxels, as illustrated in Fig.~\ref{fig:representation}(b).
As a representative example, PixelNeRF defines the output as
\begin{equation}
    (\sigma, \mat{c}) = \mathcal{F}_{\text{2D}}(\mat{W}(\pi(\mat{x})); \mat{x}_c; \mat{d}_c),
\end{equation}
where $\mat{x}_c$ is the 3D position and $\mat{d}_c$ is the ray direction.
In this case, the spatial information is encoded inside the feature map $\mat{W}$ when it is extracted by an image encoder.
Consequently, any 3D point along a ray $\mat{x}_t \in \mat{r}$ would share the same feature $\mat{W}(\pi(\mat{x}_t))$.
This representation encourages better rendering quality in visible areas, and is more computationally efficient.
However, it often generates blurry predictions in unseen parts shown in Fig.~\ref{fig:teaser}.

\boldstartspace{3D volume-based approaches.}
To utilize 3D locality, another way is to treat the object as a set of voxels in 3D space and apply 3D convolutions to reconstruct unseen areas (see Fig.~\ref{fig:representation}(c)).
The voxel grid can be constructed by unprojecting 2D images or feature maps to a 3D volume \cite{fast-and-explicit-neural-view-synthesis}.
For each 3D point, we have features $\mat{W}(\pi(\mat{x}))$ and 3D location $\mat{x}$.
The 3D CNN can utilize information from neighboring voxels to infer geometry and appearance at $\mat{x}$ as follows
\begin{equation}
    (\sigma, \mat{c}) = \mathcal{F}_{\text{3D}}(\mat{W}(\pi(\mat{x}_n));\mat{x}_n),
\end{equation}
where $\mat{x}_n$ denotes the set of neighboring voxels of $\mat{x}$.
This method is faster in rendering, and leverages 3D prior to rendering unseen geometry.
On the other hand, it suffers from limited rendering resolution due to the voxel size and limited receptive fields. 

\boldstartspace{Our approach.} We observe that the 1D approach enjoys a holistic view on the object and is able to encode the overall shape in a compact format.
The 2D method offers better visual quality around input views, while the 3D method refines the shape.
However, volume-based methods are more computationally-intensive and require more memory when increasing the grid size. 
Our method combines the advantage of 2D-based method that condition on local image features, and 1D-based methods that encode global information.
Specifically, we utilize (i) a ViT architecture and its fully-connected networks to learn global information, and (ii) a 2D CNN module to extract local image features.
Recent success in vision transformer~\cite{dosovitskiy2020vit,Ranftl2021} shows the efficacy of using ViT to learn the long-range dependencies between features. 
Thus, our local and global hybrid representation allows for more flexibility and better reconstruction quality in the unseen regions.
Unlike CodeNeRF~\cite{jang2021codenerf} and DISN~\cite{NIPS2019_8340}, our method does not require a canonical coordinate system to utilize the global features.
Our method enjoys the benefits of  high-resolution image features from 2D-CNN, while improving the receptive fields through ViT encoder.

%


\subsection{Global Features from Vision Transformer}\label{sec:vit}

We adopt the image-based approach that conditions on per-pixel feature $\mat{W}$ for rendering.
We divide $\mat{W}$ into two parts: (i) global feature maps $\mat{W}_G$ and (ii) local feature maps $\mat{W}_L$.
In this section, we describe how we obtain $\mat{W}_G$ with a vision transformer.
Our model takes as an input a single image $\mat{I}_s \in \mathbb{R}^{H\times W \times 3}$, where $H$ and $W$ are the image height and width, respectively.

\boldstartspace{Flatten and project.} As shown in Fig.~\ref{fig:algorithm}, the image $\mat{I}_s$ is first reshaped into a sequence of flattened 2D patches $\mat{P} \in \mathbb{R}^{N\times P^2 \times 3}$, where $N=\tfrac{HW}{P^2}$ is the number of patches, and $P$ denotes the patch size~\cite{dosovitskiy2020vit}.
As the transformer takes a latent vector of size $D$, we project the patches with a trainable linear layer to produce $\mat{P}_l \in \mathbb{R}^{N\times D}$.
In previous ViT work~\cite{dosovitskiy2020vit}, 
a learnable class token is usually concatenated to the image tokens to incorporate global information that is not grounded in the input image.
In our case, we treat the class token as a ``background" token to represent features that are not shown in the image.
Consequently, we have $N+1$ tokens in total, denoted as $\mat{P}_l^0, \mat{P}_l^1, ..., \mat{P}_l^{N}$.
We also add learnable positional embeddings $\mat{e}$ to distinguish between different spatial patches:
$\mat{P}_e^i = \mat{P}_l^i + \mat{e}^i.$

\boldstartspace{Transformer encoder.} The tokens $\{\mat{P}_e^0, \mat{P}_e^1, ..., \mat{P}_e^{N}\}$ undergo $J$ transformer layers to generate latent features $f^j$, where $j$ denotes the output of the $j$-th transformer layer.
The transformer layer is composed of multiheaded self-attention (MSA) and MLP layers~\cite{dosovitskiy2020vit}.
The MSA block performs self-attention on the images and extracts information by comparing a pair of tokens.
Therefore, the transformer encoder has a global receptive field in all the layers, which can easily learn long-range dependency between different image patches~\cite{dosovitskiy2020vit,Ranftl2021}.
%
%

\boldstartspace{Convolutional decoder.} After generating a set of latent features $f=\{f^0, ...,\allowbreak f^J\}, f^j \in \mathbb{R}^D$, our algorithm then utilizes a convolutional decoder to promote the latent features into multi-level feature maps.
These multi-level feature maps extract coarse-to-fine global information and allow us to concatenate with the local appearance features in the final rendering stage (see Sec.~\ref{sec:cnn}).
%
To generate the feature maps, we first drop the class token.
The class token is useful during the self-attention stage but does not have physical meaning when unflattened~\cite{Ranftl2021}.
Consequently, we define the operation as 
$\mathcal{O}: \mathbb{R}^{(N+1)\times D} \rightarrow \mathbb{R}^{N\times D}$.
After dropping the class token, we unflatten the image by 
$\mathcal{U}: \mathbb{R}^{N\times D} \rightarrow \mathbb{R}^{\frac{H}{P}\times \frac{W}{P} \times D}$.
Now we have a set of feature patches $\mat{P}_f=\{\mat{P}^0_f, ..., \mat{P}^J_f\}$, where $\mat{P}^j_f\in \mathbb{R}^{\frac{H}{P}\times \frac{W}{P} \times D}$.
We then construct the multi-level feature maps with a set of convolutional decoders as in Fig.~\ref{fig:algorithm}.
The convolutional decoders are defined as
$\mathcal{D}: \mathbb{R}^{\frac{H}{P}\times \frac{W}{P} \times D} \rightarrow \mathbb{R}^{H^j\times W^j \times D^j}$,
where the feature patches are (i) first convolved with a $1\times1$ convolution layer, (ii) resampled with a strided convolution or transposed convolution to have size $H^j\times W^j$, and (iii) convolved with a $3\times3$ convolution layer to have $D^j$ channels.
We can describe the feature maps as,
\begin{equation}
    \mat{W}^j_G = (\mathcal{D} \circ \mathcal{U} \circ \mathcal{O})(f^j), \text{where } j \in \{0, 1, ..., J\}.
\end{equation}

\subsection{Local Features from Convolutional Networks}\label{sec:cnn}

We empirically find that only using the global information from ViT compromises the rendering quality of target views that are close to the input view, \eg, the color and appearance are inconsistent (see Fig.~\ref{fig:ablation}).
%
To alleviate this problem, we introduce an additional 2D CNN module $\mathcal{G}_L$ to extract local image features, which can improve the color and appearance consistency in the visible regions.
The local features can be represented as
\begin{equation}
    \mat{W}_L = \mathcal{G}_L(\mat{I}_s), \mathcal{G}_L: \mathbb{R}^{H\times W\times C} \rightarrow \mathbb{R}^{\frac{H}{2} \times \frac{W}{2} \times D_L},
\end{equation}
where $D_L$ is the output dimension of $\mathcal{G}_L$.

Finally, we use a convolutional layer $\mathcal{G}$ to fuse the information from both global feature $\mat{W}_G$ and local feature $\mat{W}_L$ and generate the hybrid feature map:
\begin{equation}
    \mat{W} = \mathcal{G}(\mat{W}^0_G, \mat{W}^1_G, ..., \mat{W}^J_G; \mat{W}_L)
\end{equation}

\subsection{Volumetric Rendering with NeRF}\label{sec:volume_rendering}

Once we obtain the hybrid features $\mat{W}$, we can adopt the volumetric rendering~\cite{mildenhall2020nerf} to render a target view conditioned on $\mat{W}$.
We start by sampling a ray $\mat{r}(t) = \mat{o} + t\mat{d}$ from the target viewpoint, where $\mat{o}$ is the origin of the ray, $\mat{d}$ is the ray direction, and $t$ is the distance from the origin.
Note that $t$ is bounded by near plane $t_{\text{near}}$ and far plane $t_{\text{far}}$.
Along the ray, we first pick equally distributed samples between the bounds $[t_{\text{near}}, t_{\text{far}}]$.
We denote a 3D sample location as $\mat{x}$, which can be projected onto the source image with coordinate $\pi(\mat{x})$ with known camera parameters.
We then extract the per-pixel feature as $\mat{W}(\pi(\mat{x}))$.
The NeRF MLP module takes as input the per-pixel feature $\mat{W}(\pi(\mat{x}))$, 3D sample location in camera coordinate $\mat{x}_c$ and viewing direction $\mat{d}_c$.
We encode $\mat{x}_c$ with positional encoding $\gamma$:
\begin{equation}
\begin{split}
    \gamma(p) = (\sin(2^0\pi p), \cos(2^0\pi p), ...,\\
                \sin(2^{M-1}\pi p), \cos(2^{M-1}\pi p)),
\end{split}
\end{equation}
where $M$ is the number of frequency bases.
We set $M=10$ in all our experiments.
The MLP outputs color $\mat{c}$ and density $\sigma$, which can be written as:
\begin{equation}
    (\sigma, \mat{c}) = \text{MLP}(\gamma(\mat{x}_c); \mat{d}_c; \mat{W}(\pi(\mat{x}))).
\end{equation}
Finally, we render the target view into a 2D image via
\begin{equation}
    \hat{\mat{C}}(\mat{r}) = \int^{t_f}_{t_n}T(t)\sigma(t)\mat{c}(t)dt,
\end{equation}
where $T(t)=\exp(-\int^{t}_{t_n}\sigma(\mat{r}(s))ds)$ is the accumulated transmittance along the ray from $t_n$ to $t$.
Here we approximate the integral with quadrature~\cite{mildenhall2020nerf}.

We adopt a L2 norm loss to compare the rendered pixel $\hat{\mat{C}}(\mat{r})$ against the ground-truth pixel:
\begin{equation}
    \mathcal{L} = \sum_{\mat{r}}||\hat{\mat{C}}(\mat{r})-\mat{C}(\mat{r})||^2_2.
\end{equation}

\boldstart{Implementation details.}\label{sec:implementation}
We implement our method using PyTorch~\cite{NEURIPS2019_9015}.
The ViT module is initialized from the pretrained weights of~\cite{rw2019timm} and fine-tuned with the training.
The 2D CNN module $\mathcal{G}_L$ has three ResBlocks.
The detailed architecture of the entire model is provided in the supplementary material.
We train our model on 16 NVIDIA A100 GPUs, where the training converges at 500K iterations. 
We set the learning rate to be $10^{-4}$ for the MLP and $10^{-5}$ for ViT and the CNN.
To improve training stability, we use a linear warm-up schedule to increase the learning rate linearly from $0$ for the first 10k steps. Please see our supplementary material for more details.
As NeRF is trained with batches of rays, we use 512 rays for 1 instance and the batch size of 8.
\section{Experimental Results}

To evaluate our method, we conduct experiments on category-specific view synthesis (Sec.~\ref{sec:category_specific}) and category-agnostic view synthesis (Sec.~\ref{sec:category_agnostic}).
%
Sec.~\ref{sec:real} shows the qualitative results of our method on real input images.
Sec.~\ref{sec:ablation} provides ablation studies to analyze the key components in our method.
%
Sec.~\ref{sec:backbone} replaces the ViT with different backbones and show the efficacy of using ViT features.
Finally, we discuss the limitations and future work (Sec.~\ref{sec:limitations}).

\begin{figure}[t]
\centering
\includegraphics[width=1.0\textwidth]{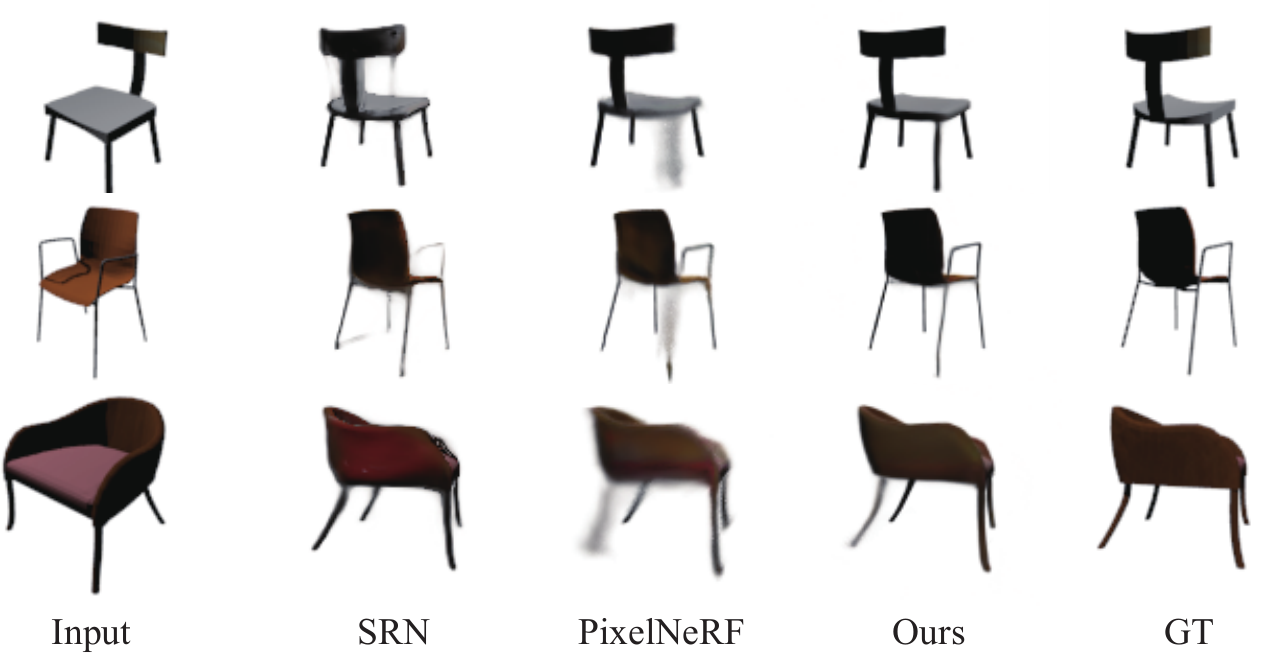}
\caption{\boldstart{Category-specific view synthesis on Chairs.} The results of SRN and PixelNeRF are often too blurry, especially on the legs that are not visible in the input views.
Our method can generate novel views with clearer structures and sharper edges.
}
\label{fig:category_specific_chairs}
\end{figure}

\begin{figure}[t]
\centering
\includegraphics[width=1.0\textwidth]{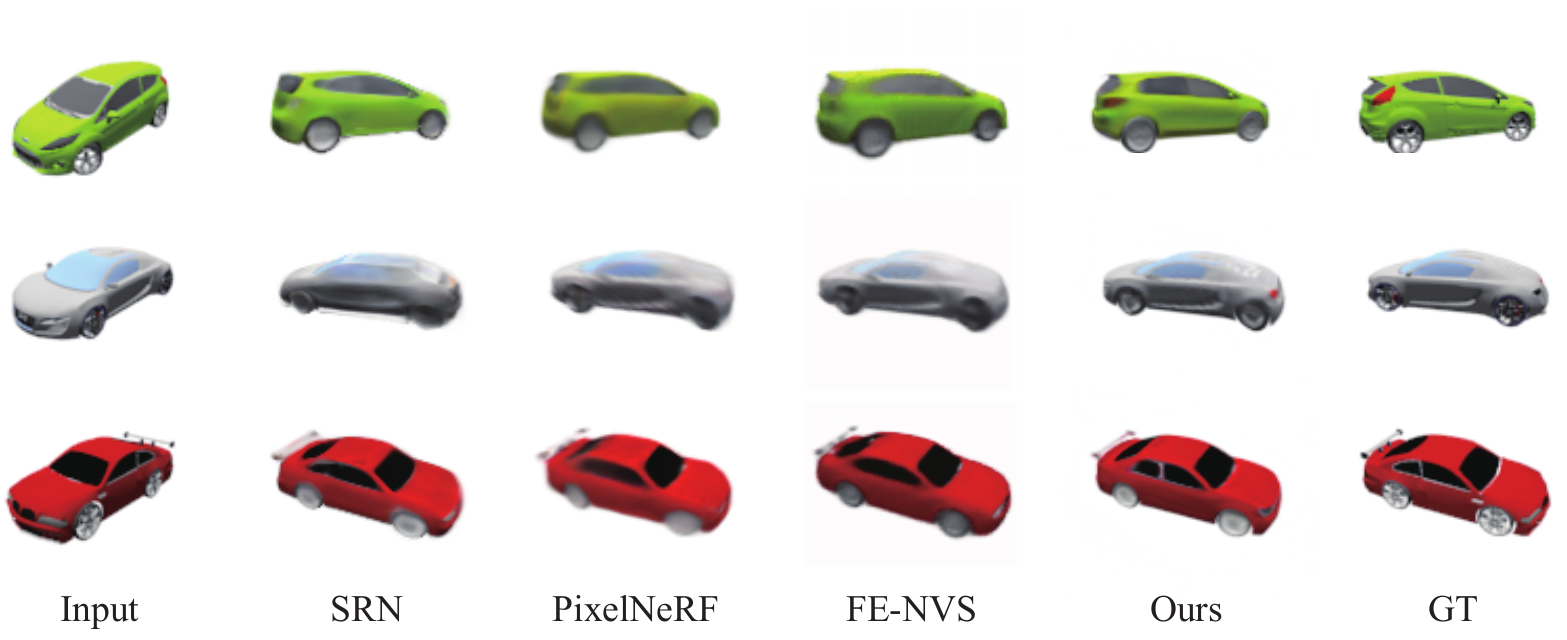}
\caption{\boldstart{Category-specific view synthesis on Cars.} Our method can generate sharper car structure and richer details, such as the rear lights and windows in the first row, the wheels and door in the second row, and the windows in the third row.
}
\label{fig:category_specific_cars}
\end{figure}


\subsection{Category-specific View Synthesis}\label{sec:category_specific}

We evaluate our method on the same experimental setup and data as SRN~\cite{sitzmann2019srns}.
The dataset consists of 6591 chairs and 3514 cars in total, which are split into training, validation, and test sets.
For each object in the training set, 50 views lying on a sphere around the object are selected to render with simple lighting.
For testing, the objects in the test set are rendered from 251 views on an archimedean spiral with the same illumination as training.
During the evaluation, the 64-th view is selected as the input view and all other 250 views are used as target views.
The image resolution is $128\times 128$.
We compare our method with SRN~\cite{sitzmann2019srns}, PixelNeRF~\cite{yu2020pixelnerf}\footnote{LPIPS is calculated from the results provided by the authors.}, CodeNeRF~\cite{jang2021codenerf}\footnote{LPIPS and code for unposed inference are not available.} and FE-NVS~\cite{fast-and-explicit-neural-view-synthesis}\footnote{LPIPS is provided by the authors on request.}.


As shown in Table~\ref{tab:category_specific}, our method achieves state-of-the-art performance against existing approaches in terms of PSNR, SSIM, and LPIPS~\cite{zhang2018perceptual}.
On the chair dataset, our method shows significant improvement on all three metrics.
As shown in Fig.~\ref{fig:category_specific_chairs}, our rendered results have better appearance and clearer structures, while SRN~\cite{sitzmann2019srns} and PixelNeRF~\cite{yu2020pixelnerf} have blurry predictions on the chair legs.
On the car dataset, we obtain the best LPIPS and SSIM scores.
While PixelNeRF~\cite{yu2020pixelnerf} has the highest PSNR, their results are overly-blurry with smooth textures, as shown in Fig.~\ref{fig:category_specific_cars}.
In contrast, our predictions have finer details and reveal more details such as the windows, lights, and wheels. 
Note that we do not compare visual results with CodeNeRF~\cite{jang2021codenerf} as their pre-generated results are not publicly available, and their source code does not support inference without camera poses.
FE-NVS~\cite{fast-and-explicit-neural-view-synthesis} does not provide source code or pre-generate results as well.
However, we try our best to obtain high-resolution screenshots from their paper and compare with their results on the same view.

%

\setlength{\tabcolsep}{0.5em} 
\begin{table}[t]
  \centering
   {\caption{\boldstart{Category-specific view synthesis on the ShapeNet dataset.} 
   Our method performs favorably against other approaches, especially on LPIPS.
   Note that while PixelNeRF has higher PSNR on the cars dataset, their results look blurry (see Fig.~\ref{fig:category_specific_cars}).
  }
  \label{tab:category_specific}}
  \resizebox{\linewidth}{!}{
    {\begin{tabular}{l c c c c c c} 
      \toprule
                            &                & Chairs         &                   &                & Cars           & \\
        Methods             & PSNR($\uparrow$) & SSIM($\uparrow$) & LPIPS($\downarrow$) & PSNR($\uparrow$) & SSIM($\uparrow$) & LPIPS($\downarrow$)\\
        \midrule
        SRN~\cite{sitzmann2019srns}                 &     22.89 &      0.89  & \cs{0.104} &     22.25  & \ct{0.89} &     0.129  \\
        PixelNeRF~\cite{yu2020pixelnerf}            & \cs{23.72} & \ct{0.91} & \ct{0.128} & \cb{23.17} & \cs{0.90} &     0.146  \\
        CodeNeRF~\cite{jang2021codenerf}            &     22.39  &     0.87  &     0.166  &     22.73  & \ct{0.89} & \ct{0.128} \\
        FE-NVS~\cite{
        fast-and-explicit-neural-view-synthesis}    & \ct{23.21} & \cs{0.92} & \cb{0.077} & \ct{22.83} & \cb{0.91} & \cs{0.099} \\
        \textbf{Ours}                               & \cb{24.48} & \cb{0.93} & \cb{0.077} & \cs{22.88} & \cb{0.91} & \cb{0.084} \\
        \bottomrule
      \end{tabular}}
    }
\end{table}

\begin{table*}[t]
  {\caption{\boldstart{Category-agnostic view synthesis on the NMR dataset.} 
  Our method achieves the state-of-the-art performance across all 13 categories using a single model.
  }\label{tab:category_agnostic}}
  \begin{adjustbox}{width=0.95\columnwidth,center}
      \begin{tabular}{l l c c c c c c c c c c c c c c}
      \toprule
        Metrics & Methods & plane & bench & cbnt. & car & chair & disp. & lamp & spkr. & rifle & sofa & table & phone & boat & average \\
        \midrule
        \multirow{4}{*}{PSNR($\uparrow$)}    & SRN       & 26.62 & 22.20 & 23.42 & 24.40 & 21.85 & 19.07 & 22.17 & 21.04 & 24.95 & 23.65 & 22.45 & 20.87 & 25.86 & 23.28 \\
                                           & PixelNeRF & 29.76 & 26.35 & 27.72 & 27.58 & 23.84 & 24.22 & \cs{28.58} & 24.44 & \cs{30.60} & 26.94 & 25.59 & 27.13 & \cs{29.18} & 26.80 \\
                                           & FE-NVS    & \ct{30.15} & \ct{27.01} & \ct{28.77} & \ct{27.74} & \ct{24.13} & \ct{24.13} & 28.19 & \ct{24.85} & 30.23 & \ct{27.32} & \ct{26.18} & 27.25 & \ct{28.91} & \ct{27.08} \\
                                           & FWD       & 30.01 & 26.16 & 28.49 & 27.01 & 23.44 & 24.00 & 27.84 & 24.45 & \ct{30.40} & 26.76 & 25.91 & \ct{27.61} & 28.69 & 26.66 \\
                                           & SRT       & \cs{31.47} & \cs{28.45} & \cs{30.40} & \cs{28.21} & \cs{24.69} & \cs{24.58} & \ct{28.56} & \cs{25.61} & 30.09 & \cs{28.11} & \cs{27.42} & \cs{28.28} & \cs{29.18} & \cs{27.87} \\
                                         &\textbf{Ours}& \cb{32.34} & \cb{29.15} & \cb{31.01} & \cb{29.51} & \cb{25.41} & \cb{25.77} & \cb{29.41} & \cb{26.09} & \cb{31.83} & \cb{28.89} & \cb{27.96} & \cb{29.21} & \cb{30.31} & \cb{28.76} \\
        \midrule
        \multirow{4}{*}{SSIM($\uparrow$)}    & SRN       & 0.901 & 0.837 & 0.831 & 0.897 & 0.814 & 0.744 & 0.801 & 0.779 & 0.913 & 0.851 & 0.828 & 0.811 & 0.898 & 0.849 \\
                                           & PixelNeRF & 0.947 & 0.911 & 0.910 & \ct{0.942} & 0.858 & \ct{0.867} & \ct{0.913} & 0.855 & \ct{0.968} & 0.908 & 0.898 & 0.922 & 
                                           \ct{0.939} & 0.910 \\
                                           & FE-NVS    & \cs{0.957} & \cs{0.930} & \cs{0.925} & \cs{0.948} & \cs{0.877} & \cs{0.871} & \cs{0.916} & \cs{0.869} & \cs{0.970} & \cs{0.920} & \cs{0.914} & \cs{0.926} & \cs{0.941} & \cs{0.920} \\
                                           & FWD       & 0.952 & 0.914 & 0.918 & 0.939 & 0.857 & \ct{0.867} & 0.906 & \ct{0.857} & \ct{0.968} & 0.909 & 0.906 & 
                                           \ct{0.924} & 0.936 & 0.911 \\
                                           & SRT       & \ct{0.954} & \ct{0.925} & \ct{0.920} & 0.937 & \ct{0.861} & 0.855 & 0.904 & 0.854 & 0.962 & \ct{0.911} & \ct{0.909} & 0.918 & 0.930 & \ct{0.912} \\
                                         &\textbf{Ours}& \cb{0.965} & \cb{0.944} & \cb{0.937} & \cb{0.958} & \cb{0.892} & \cb{0.891} & \cb{0.925} & \cb{0.877} & \cb{0.974} & \cb{0.930} & \cb{0.929} & \cb{0.936} & \cb{0.950} & \cb{0.933} \\
        \midrule
        \multirow{4}{*}{LPIPS($\downarrow$)} & SRN       & 0.111 & 0.150 & 0.147 & 0.115 & 0.152 & 0.197 & 0.210 & 0.178 & 0.111 & 0.129 & 0.135 & 0.165 & 0.134 & 0.139 \\
                                           & PixelNeRF & 0.084 & 0.116 & 0.105 & 0.095 & 0.146 & 0.129 & 0.114 & 0.141 & 0.066 & 0.116 & 0.098 & 0.097 & 0.111 & 0.108 \\
                                           & FE-NVS    & 0.061 & 0.080 & 0.076 & 0.085 & 0.103 & 0.105 & 0.091 & 0.116 & 0.048 & 0.081 & \ct{0.071} & 0.080 & 0.094 & 0.082 \\
                                           & FWD       & \cb{0.034} & \cb{0.055} & \cb{0.056} & \cb{0.042} & \cb{0.081} & \cb{0.079} & \cb{0.062} & \cb{0.091} & \cb{0.026} & \cb{0.054} & \cb{0.049} & \cb{0.056} & \cb{0.052} & \cb{0.055} \\
                                           & SRT       & \ct{0.050} & \ct{0.068} & \cs{0.058} & \ct{0.062} & \ct{0.085} & \ct{0.087} & \ct{0.082} & \cs{0.096} & \cs{0.045} & \cs{0.066} & \cs{0.055} & \cs{0.059} & \ct{0.079} & \ct{0.066} \\
                                         &\textbf{Ours}& \cs{0.042} & \cs{0.067} & \ct{0.065} & \cs{0.059} & \cs{0.084} & \cs{0.086} & \cs{0.073} & \ct{0.103} & \ct{0.046} &   \ct{0.068} & \cs{0.055} & \ct{0.068} & \cs{0.072} & \cs{0.065} \\
        \bottomrule
      \end{tabular}
  \end{adjustbox}
\end{table*}

\begin{figure}[t]
\centering
\includegraphics[width=1.0\textwidth]{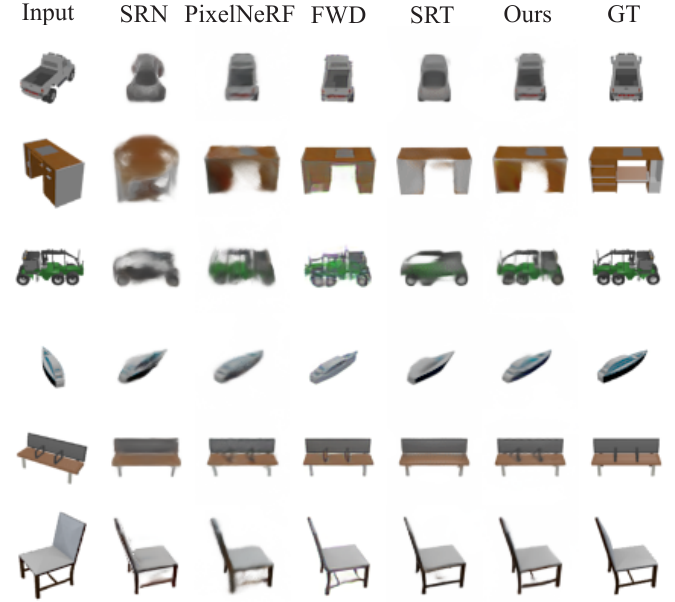}
\caption{\boldstart{Visual comparison of category-agnostic view synthesis.} 
The results of SRN~\cite{sitzmann2019srns}, PixelNeRF~\cite{yu2020pixelnerf} and SRT~\cite{srt22} are often too blurry and contain smearing artifacts.
In contrast, our results are sharper with more fine details.
FWD~\cite{Cao2022FWD} produces distorted renderings at far viewpoints because the depth is not as accurate for occluded regions.
The visual results of all 13 categories are provided in the supplementary material. 
}
\label{fig:category_agnostic}
\end{figure}

\begin{figure}[t]
\centering
\includegraphics[width=1.0\textwidth]{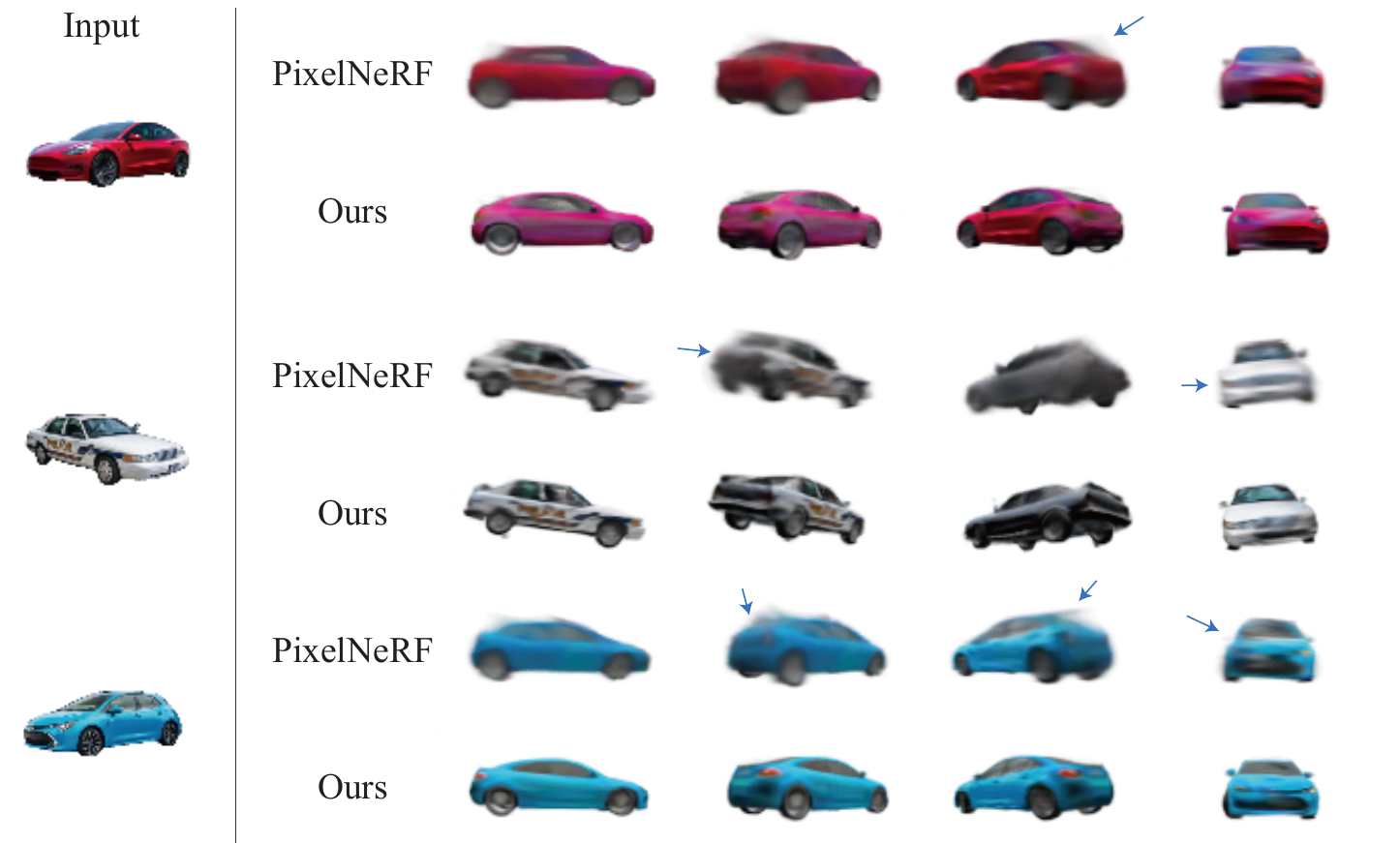}
\caption{\boldstart{Results on real input images.} Our method is able to generate visually-pleasing results even trained on a synthetic dataset. Conversely, PixelNeRF fails to keep the finer details. Note the side mirrors and headlamps of the bottom right inset. 
}
\label{fig:real}
\end{figure}

\begin{figure}[h!]
\centering
\includegraphics[width=1.0\textwidth]{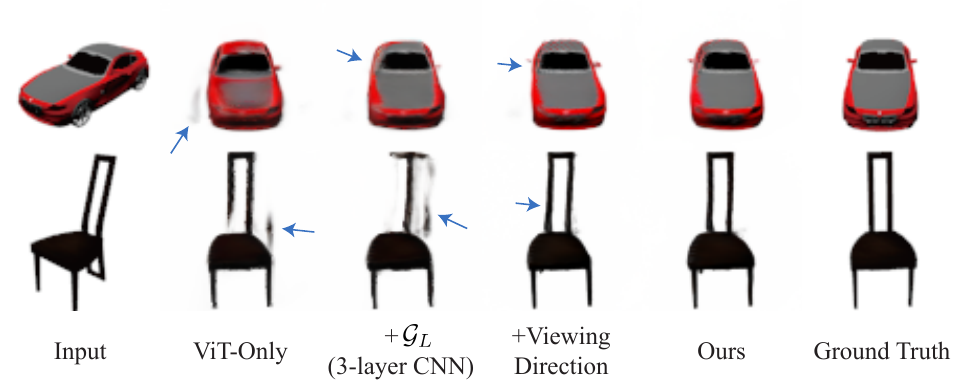}
\caption{\boldstart{Effects of different components.} 
    The ViT-only model can render realistic images, but the local appearance and color may not look similar to the input view.
    By extracting local features with a 3-layer CNN, the rendered car shows more faithful colors to the input.
    With the viewing direction in volume rendering, our model can improve fine structures such as the left mirror of the car and the back of the chair.
    In our final model, replacing the 3-layer CNN with ResBlocks can further refine the details and geometry structure of the rendered objects. 
}
\label{fig:ablation}
\end{figure}

\begin{table}[t]
  \centering
  {\caption{\boldstart{Ablation studies.} 
    We start from a baseline model that uses ViT to extract global features.
    While PSNR/SSIM are slightly lower than PixelNeRF, our results have much better LPIPS scores and sharper details (see Fig.~\ref{fig:teaser}).
    By using a 3-layer CNN to extract local features, our performance on the car dataset is improved, and the rendered images have more faithful appearances to the input views (see Fig.~\ref{fig:ablation}).
    By adding the viewing direction in volume rendering, the performance is improved significantly.
    Finally, by replacing the 3-layer CNN with ResBlocks, we see more fine details and better object structure in Fig.~\ref{fig:ablation}.}
  \resizebox{\linewidth}{!}{
  {\begin{tabular}{@{\extracolsep{1pt}}l c c c c c c@{}}
  \toprule
                                &                &    Cars      &                   &                &      Chairs      &                   \\
    Method                      & PSNR$\uparrow$ & SSIM$\uparrow$ & LPIPS$\downarrow$ & PSNR$\uparrow$ & SSIM$\uparrow$ & LPIPS$\downarrow$ \\
    \midrule
    PixelNeRF                       & 23.17 & 0.90 & 0.146 & 23.72 & 0.91 & 0.128 \\
    \midrule
    ViT only                        &      21.95 & \ct{0.89} &     0.130  & \ct{23.45} & \cs{0.92} &     0.099 \\
    + $\mathcal{G}_L$ (3-layer CNN) & \ct{22.42} & \cs{0.90} & \ct{0.113} & 23.42      & \cs{0.92} & \cs{0.085} \\
    + Viewing Direction             & \cs{22.70} & \cb{0.91} & \cs{0.088} & \cb{24.53} & \cb{0.93} & \ct{0.094} \\
    \textbf{Ours}   & \cb{22.88} & \cb{0.91} & \cb{0.084} & \cs{24.48} & \cb{0.93} & \cb{0.077} \\
    \bottomrule
  \end{tabular}}
  }
    \label{tab:ablation}}
\end{table}

\subsection{Category-agnostic View Synthesis}\label{sec:category_agnostic}
Our method is able to generalize across multiple object categories using a single model.
We follow the training/test splits of the ShapeNet dataset defined in NMR~\cite{kato2018renderer} and choose 1 view as input while the other 23 views as target in both training and evaluation.
There are 30642 objects for training and 8762 objects for evaluation (from 13 categories).
The image resolution is $64\times 64$.


Table~\ref{tab:category_agnostic} shows the quantitative results. Our method achieves the state-of-the-art performance against SRN~\cite{sitzmann2019srns}, PixelNeRF~\cite{yu2020pixelnerf}, FE-NVS~\cite{fast-and-explicit-neural-view-synthesis}, FWD~\cite{Cao2022FWD} and SRT~\cite{srt22} on all 13 categories in PSNR and SSIM.
Our method achieves competitive performance in LPIPS compared to recent state-of-the-art approaches, FWD~\cite{Cao2022FWD} and SRT~\cite{srt22}.
The results demonstrate that our hybrid representation is more expressive than the locally-conditioned models or 3D voxel methods.
The visual comparisons in Fig.~\ref{fig:category_agnostic} shows that our method reconstructs finer object structure and details.
Even though FWD~\cite{Cao2022FWD} achieves better LPIPS scores, their results show distorted renderings at larger displacements, possibly due to erroneous depth estimation at unseen areas. 
In Fig.~\ref{fig:category_agnostic}, the vehicle on the third row shows severe artifacts when FWD tries to render viewpoints at the opposite side of the input. Please refer to supplementary materials for more examples.
Since SRT~\cite{srt22} converts input images to a set of latent codes without a one-to-one mapping to the source image, their results often lose fine details, \eg, the bench on the second to the last row in Fig.~\ref{fig:category_agnostic}.

\subsection{View Synthesis on Real Images}\label{sec:real}

Our method generalizes to real images.
We use our model trained on the ShapeNet car dataset to test on real car images from the Stanford cars dataset~\cite{KrauseStarkDengFei-Fei_3DRR2013}.
%
We use an image segmentation model~\cite{kirillov2019pointrend} to remove the background. 
Note that our method does not require any camera pose as input, which is often difficult to obtain from real images.
We compare our results with PixelNeRF in Fig.~\ref{fig:real}.
%
In the occluded regions, PixelNeRF suffers from blurry predictions as pointed out by the arrows.
In contrast, our method is able to reconstruct the entire shape and keep details such as headlights and side mirrors.

\begin{table}[t]
  \centering
  {\caption{
  \boldstart{Comparison with different backbone choices.}
  %
  We replace the vision transformer with EfficientNet~\cite{tan2019efficientnet} and ConvNeXt~\cite{liu2022convnet} to observe potential performance impact.
  Our method achieves favorable overall performance in LPIPS compared to other backbones.
  }
  \resizebox{\linewidth}{!}{
  {\begin{tabular}{@{\extracolsep{1pt}}l c c c c c c@{}}
  \toprule
                                    &                &    Cars      &                   &                &      Chairs      &                   \\
    Method                          & PSNR$\uparrow$ & SSIM$\uparrow$ & LPIPS$\downarrow$ & PSNR$\uparrow$ & SSIM$\uparrow$ & LPIPS$\downarrow$ \\
    \midrule
    PixelNeRF                       & 23.17 & 0.90 & 0.146 & 23.72 & 0.91 & 0.128 \\
    \midrule
    Replace ViT with EfficientNet   & \cs{23.28} & 0.91 & \ct{0.106} & \ct{24.09} & \cs{0.92} & \ct{0.105} \\
    Replace ViT with ConvNeXt       & \cb{23.30} & 0.91 & \cs{0.092} & \cs{24.37} & \cb{0.93} & \cs{0.089} \\
    \textbf{Ours}                   & \ct{22.88} & 0.91 & \cb{0.084} & \cb{24.48} & \cb{0.93} & \cb{0.077} \\
    \bottomrule
  \end{tabular}}
  }
    \label{tab:backbone}}
\end{table}

\subsection{Ablation Studies}\label{sec:ablation}
%
We start from the baseline method using only the ViT to extract global features.
While ViT encodes the high-level global information, it fails to preserve the color and appearance from the input view due to the low-resolution latent embeddings, as shown in Fig.~\ref{fig:ablation}.
The rendered results show inconsistent appearances to the input view on non-occluded regions, as shown in the second column in .
By introducing $G_L$ (using a simple 3-layer CNN) to extract local image features, the rendered car looks closer to the input view (top of the third column in Fig.~\ref{fig:ablation}).
However, we can see that the chair's back is still blurry (bottom of the third column in Fig.~\ref{fig:ablation}).
Next, we add the viewing direction as input to the NeRF MLP, which significantly improves the sharpness (bottom of the 4-th column in Fig.~\ref{fig:ablation}) and reveals more details such as the rear mirror of the car (top of the 4-th column in Fig.~\ref{fig:ablation}).
Our final model adopts a more complex ResBlocks design in $G_L$, which further improves the geometry shape of the car and chair (the 5-th column in Fig.~\ref{fig:ablation}).
Table~\ref{tab:ablation} also reports the quantitative results of these design decisions on both datasets.
%


\subsection{Global Features from Different Backbones}\label{sec:backbone}
To further verify that ViT outperforms convolutional backbones for image-conditioned NeRFs, we benchmark our method against two baselines that replace the proposed ViT backbone with EfficientNet~\cite{tan2019efficientnet} and ConvNeXt~\cite{liu2022convnet}, i.e., modern CNN models with better performance than ResNet34 and comparable numbers of parameters to ViT.
The results are presented in Table~\ref{tab:backbone} which shows that our method achieves better LPIPS compared to these baselines on both the car and chair categories.
This ablation study demonstrates that using ViT as the backbone achieves better performance for image-conditioned NeRFs due to the model architecture design instead of more parameters. 

\subsection{Limitations and Future Work}\label{sec:limitations}
First, our method does not utilize geometry priors such as symmetry~\cite{wu2020unsup3d}.
For example, in the car dataset, some details on the car are symmetrical and can be reused for the unseen side.
However, it remains a question on how to select the symmetry plane or find the canonical space for such a prior.
Another limitation is that we do not fully utilize the high-level semantics of the objects.
A semantic understanding on the smaller components could help reconstruct the unseen areas much better.
For example, a car has four wheels. Given the partial observation, it is possible to use semantic knowledge to recover the unseen components.
Lastly, generative methods can be helpful in generating texture in occluded parts of the object.
Integrating locally-conditioned models with GAN loss training remains a challenging problem for future research.
\section{Conclusions}

In this work, we present a NeRF-based algorithm for novel view synthesis from a single unposed image.
We utilize vision transformer in conjunction with convolutional networks to extract global and local features as 3D representations.
This hybrid representation shows promising performance in recovering the occluded shapes and appearance.
Additionally, we show that vision transformer can be used to generate global information without enforcing a canonical coordinate system (which requires camera pose estimation).
We believe that our work has shed light on future research to synthesize faithful 3D content with local and global image features, and we hope that it could bring about more exciting advances in the frontier for immersive 3D content.
\section{Acknowledgement}
This work was supported in part by a Qualcomm FMA Fellowship, ONR grant N000142012529, ONR grant N000141912293, NSF grant 1730158, a grant from Google, and an Amazon Research award.
We also acknowledge gifts from Adobe, Google, Amazon, a Sony Research Award,
the Ronald L. Graham Chair, and the UC San Diego Center for Visual Computing.

\clearpage
\setcounter{section}{0}
\renewcommand\thesection{\Alph{section}}
\section*{Supplementary Material}
\section{Video Results}

We include video results as a webpage (\url{https://cseweb.ucsd.edu/~viscomp/projects/VisionNeRF/supplementary.html}).
%
%
In each video, we compare the input, SRN~\cite{sitzmann2019srns}, PixelNeRF~\cite{yu2020pixelnerf}, ours and ground-truth.
The renderings of SRN and PixelNeRF are provided by the authors of PixelNeRF on request.

\section{Network Architecture}

\boldstartspace{Transformer encoder.}
We adopt the pretrained ViT-b16 model from Wightman~\cite{rw2019timm} as our transformer encoder.
The transformer encoder has 12 layers ($J$=12), where each layer uses the LayerNorm~\cite{ba2016layer} for normalization and GELU~\cite{hendrycks2016gaussian} for activation.
The positional embedding is resized to the same size as the input image (e.g., $128 \times 128$ for the ShapeNet dataset, and $64 \times 64$ for the NMR dataset).  
%


\boldstartspace{Convolutional decoder.}
We provide the architecture details of our convolutional decoder in Table~\ref{tab:cnn_decoder}.
Our convolutional decoder takes the tokens $f^3$, $f^6$, $f^9$, and $f^{12}$~\cite{Ranftl2021} 
as input and generate multi-level features $\mat{W}^3_G, \mat{W}^6_G, \mat{W}^9_G,\allowbreak \mat{W}^{12}_G$, as shown in Fig.~4 of the main paper.
We then resize all feature maps $\mat{W}_G$ to the size of $\frac{H}{2}\times\frac{W}{2}$ via bilinear interpolation.
Finally, we adopt a two-layer CNN module (see Table~\ref{tab:cnn_decoder_output}) to generate the global feature representation $\mat{W}_G'$.

\setlength{\tabcolsep}{0.5em} 
\begin{table*}[h!]
  \centering
  \scriptsize
  \caption{\boldstart{Architecture of the convolutional decoder.}
  $\bf{Conv}$ denotes convolution layer.
  $\bf{TransConv}$ denotes transposed convolution layer.
  }
  \begin{tabular}{c l c c c c c c l}
    \toprule
                     $\#j$ &   layer  & kernel & stride & dilation & in & out & activation &  input \\
        \midrule
        \multirow{3}{*}{3} & Conv0-0  & $1\times1$ &      1 &        1 &768 &  96 &        N/A &  $f^3$ \\
                           & TransConv0-1  & $4\times4$ &      4 &        1 & 96 &  96 &        N/A &  Conv0-0 \\
                           & Conv0-2  & $3\times3$ &      1 &        1 & 96 & 512 &        N/A &  TransConv0-1 \\
        \midrule
        \multirow{3}{*}{6} & Conv1-0  & $1\times1$ &      1 &        1 &768 & 192 &        N/A &  $f^6$ \\
                           & TransConv1-1  & $2\times2$ &      2 &        1 &192 & 192 &        N/A &  Conv1-0 \\
                           & Conv1-2  & $3\times3$ &      1 &        1 &192 & 512 &        N/A &  TransConv1-1 \\
        \midrule
        \multirow{3}{*}{9} & Conv2-0  & $1\times1$ &      1 &        1 &768 & 384 &        N/A &  $f^9$ \\
                           & Conv2-1  & $3\times3$ &      1 &        1 &384 & 512 &        N/A &  Conv2-0 \\
        \midrule
        \multirow{3}{*}{12}& Conv3-0  & $1\times1$ &      1 &        1 &768 & 768 &        N/A & $f^{12}$ \\
                           & Conv3-1  & $3\times3$ &      2 &        1 &768 & 768 &        N/A &  Conv3-0 \\
                           & Conv3-2  & $3\times3$ &      1 &        1 &768 & 512 &        N/A &  Conv3-1 \\
    \bottomrule
  \end{tabular}
  \label{tab:cnn_decoder}
\end{table*}

\setlength{\tabcolsep}{0.5em} 
\begin{table}[h!]
  \centering
  \scriptsize
  \caption{\boldstart{Architecture of the last two layers for generating global feature representation $\mat{W}_G'$.}}
  \begin{tabular}{c c c c c c c c}
    \toprule
        Layer  & kernel & stride & dilation & in & out & activation &  input \\
    \midrule
        Conv0  & $3\times3$ &      1 &        1 &1024& 512 &       ReLU &  $\mat{W}_G$ \\
        Conv1  & $3\times3$ &      1 &        1 & 512& 256 &       ReLU &  conv0 \\
    \bottomrule
  \end{tabular}
  \label{tab:cnn_decoder_output}
\end{table}

\begin{figure}[!h]
\centering
\includegraphics[width=0.35\textwidth]{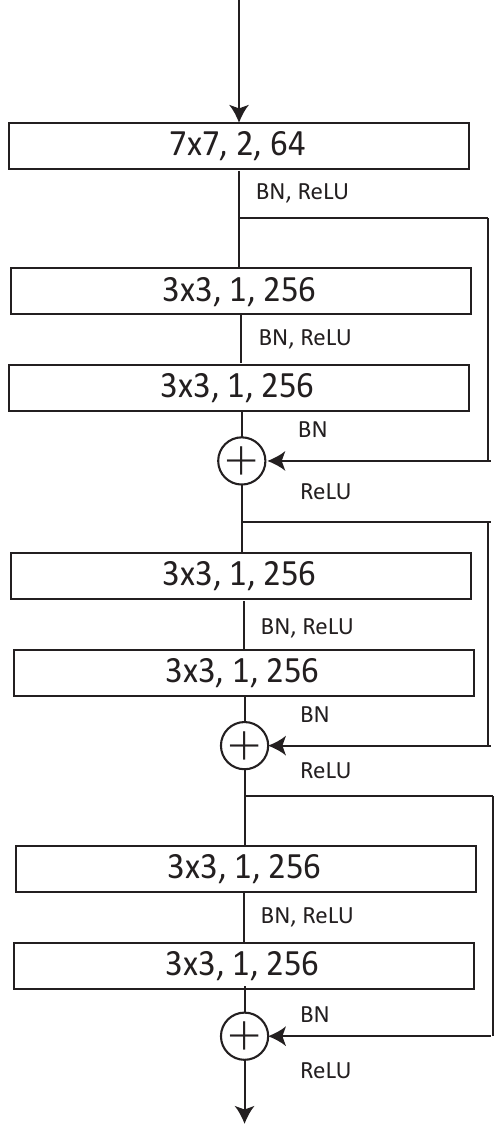}
\caption{\boldstart{Illustration of the ResBlocks.}
We use three ResBlocks as our 2D CNN module to extract local feature representation from the input image.
%
The numbers in each layer denote the kernel size, stride, and output channels, respectively.
BN denotes BatchNorm, and ReLU is the nonlinear activation function.
}
\label{fig:resnet}
\end{figure}

\boldstartspace{2D CNN $\mathcal{G}_L$.} 
We use the ResBlocks in Fig.~\ref{fig:resnet} to generate the local feature representation $\mat{W}_L$.
The local feature $\mat{W}_L$ and global feature $\mat{W}_G'$ are then concatenated as our hybrid feature representation $\mat{W}$.


\boldstartspace{NeRF MLP.}
We utilize the hierarchical rendering~\cite{mildenhall2020nerf} to include detailed geometry and appearance.
We use 6 ResNet blocks with width 512 for both coarse and fine stages to process the global and local features.


\section{Implementation Details}\label{sec:supp_implementation}

\begin{figure}
\centering
\includegraphics[width=0.9\textwidth]{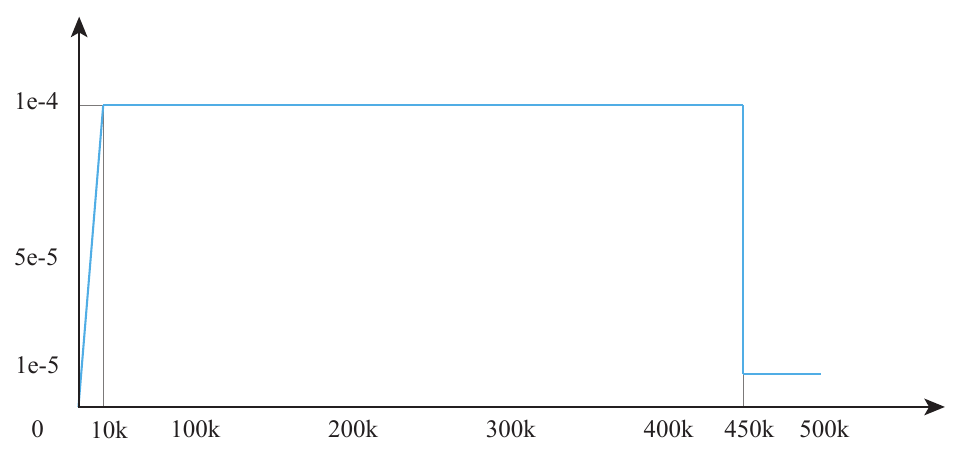}
\caption{\boldstart{Illustration of the learning rate strategy for the MLP module.}
We first linearly increase the learning rate to $1e-4$ for the first 10k step.
Then, we set a learning rate decay of scale 0.1 at 450k steps.
}
\label{fig:lr}
\end{figure}

\boldstartspace{Learning rate.}
We set the initial learning rate to be $10^{-4}$ for the MLP and $10^{-5}$ for ViT and the 2D CNN module.
To improve training stability, we use a warm-up schedule to increase the learning rate linearly from $0$ for the first 10k steps.
We decay the learning rate by a factor of $0.1$ at 450k steps. 
Our learning rate schedule for the NeRF MLP is plotted in Fig.~\ref{fig:lr}. 

\section{Experiment Configurations}
We provide the detailed experiment setups in Sec.~4 of the main paper.

\subsection{Category-specific View Synthesis.}

\boldstartspace{SRN~\cite{sitzmann2019srns} and PixelNeRF~\cite{yu2020pixelnerf}:} 
We obtain the PSNR and SSIM from Table 2 in~\cite{yu2020pixelnerf}. We use the pre-generated results provided by the authors of~\cite{yu2020pixelnerf} (on request) to calculate the LPIPS score. 

\boldstartspace{CodeNeRF~\cite{jang2021codenerf}:} 
We obtain the PSNR and SSIM of the unposed input from Table 2 in~\cite{jang2021codenerf}. As the pre-generated results are not available, and the source code of CodeNeRF does not provide optimization stages for unposed inputs, we are not able to calculate the LPIPS score.

\boldstartspace{FE-NVS~\cite{fast-and-explicit-neural-view-synthesis}:}
We obtain the PSNR and SSIM of the unposed input from Table 1 in~\cite{fast-and-explicit-neural-view-synthesis}. We obtain the LPIPS score from the authors on request.
As FE-NVS does not provide source code to reproduce any qualitative results, we crop the high-resolution images from their paper to show the comparison in Fig.~6 of the main paper.


Note that all the methods except CodeNeRF uses view 64 as input, while CodeNeRF uses view 82 as input for evaluation in their paper.
As the source code for CodeNeRF does not include unposed inference, we are not able to generate the full evaluation using view 64.

\subsection{Category-agnostic View Synthesis}

Similar to category-specific view synthesis, we obtain the numbers of SRN~\cite{sitzmann2019srns} and PixelNeRF~\cite{yu2020pixelnerf} from Table 4 of~\cite{yu2020pixelnerf}.
We obtain the numbers of FE-NVS~\cite{fast-and-explicit-neural-view-synthesis} from Table 4 of~\cite{fast-and-explicit-neural-view-synthesis}.
Qualitative results for SRN and PixelNeRF are generated using the pre-generated results from~\cite{yu2020pixelnerf}.
%
%
The detailed categorical numbers of SRT~\cite{srt22} and FWD~\cite{Cao2022FWD} are provided by their authors.
For qualitative results, we obtain the images from SRT project website and from the author of FWD.
Note that SRT does not provide the full renderings of the dataset.

\subsection{View Synthesis on Real Images}
We use the real car images provided by PixelNeRF and the Stanford Cars dataset~\cite{KrauseStarkDengFei-Fei_3DRR2013} for evaluation.
We remove the background using the PointRend~\cite{kirillov2019pointrend} segmentation and resize the images to $128\times128$ resolution.
We assume the input camera pose is an identity matrix and apply rotation matrices to simulate 360$^\circ$ views.
We use the source code and pre-trained model of PixelNeRF to generate the results for PixelNeRF, where we synthesize the renderings at the same camera poses.
\section{Additional Results}

\begin{figure}[t]
\centering
\includegraphics[width=0.75\textwidth]{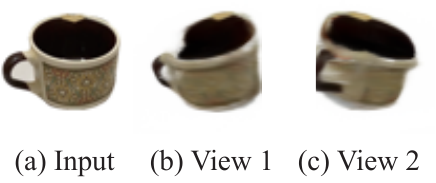}
\caption{\boldstart{Results for unseen real data.}
We run our category-agnostic model on a real image (a) and render two viewpoints from the right (b) and left (c). Our method is able to predict a reasonable geometry even though mugs are not presented in the training data.
}
\label{fig:mug}
\end{figure}

\subsection{Generalization on Unseen objects.}
For unseen categories, we feed an image of a real mug to the category-agnostic model as shown in Fig.~\ref{fig:mug}.
While the training data do not include mugs, our method is able to predict reasonable novel views.

\subsection{Results on Category-specific and Category-agnostic View Synthesis}
\begin{figure*}[t]
\centering
\includegraphics[width=0.75\textwidth]{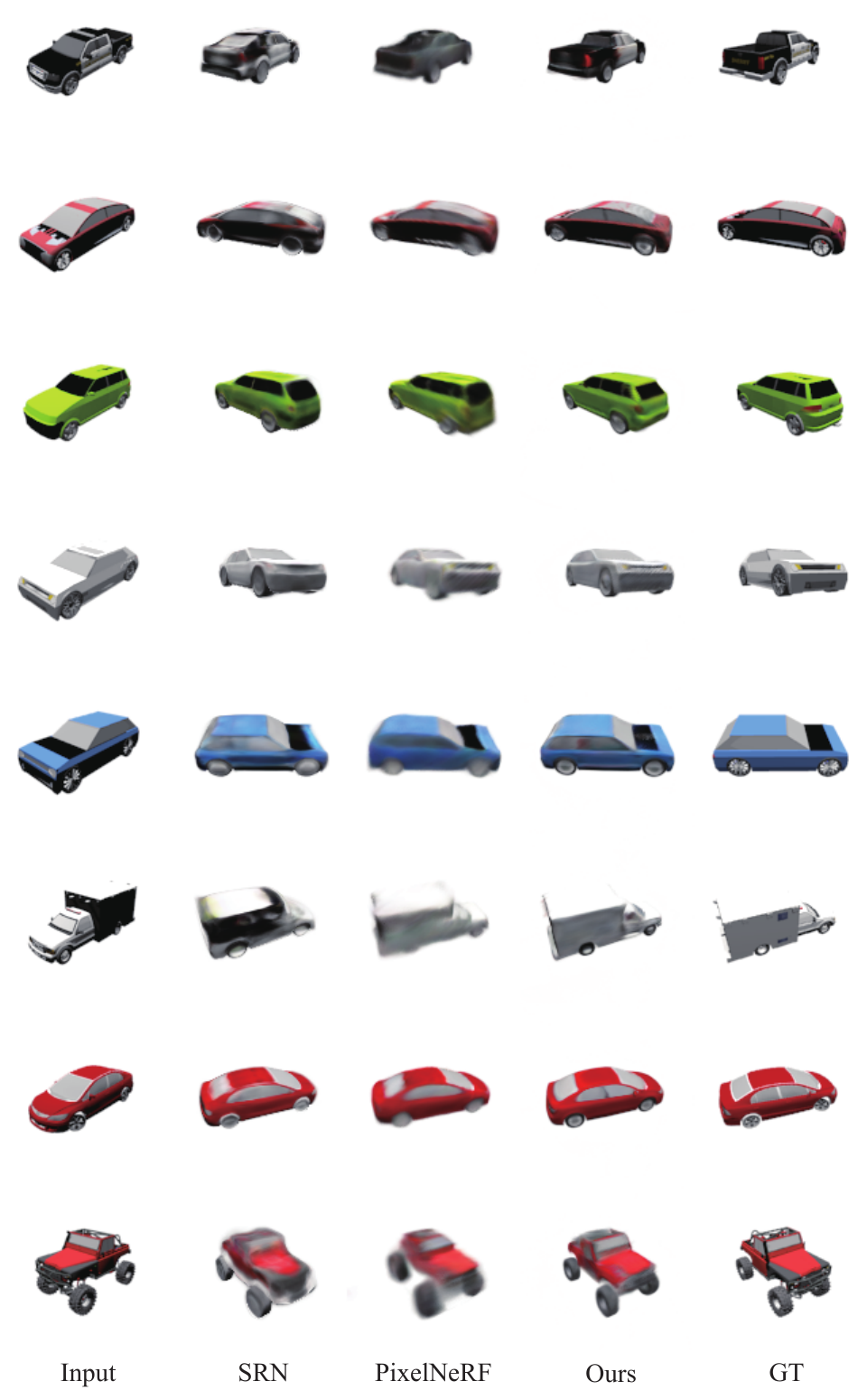}
\caption{\boldstart{Category-specific view synthesis results on cars.}}
\label{fig:cars}
\end{figure*}

\begin{figure*}[t]
\centering
\includegraphics[width=0.75\textwidth]{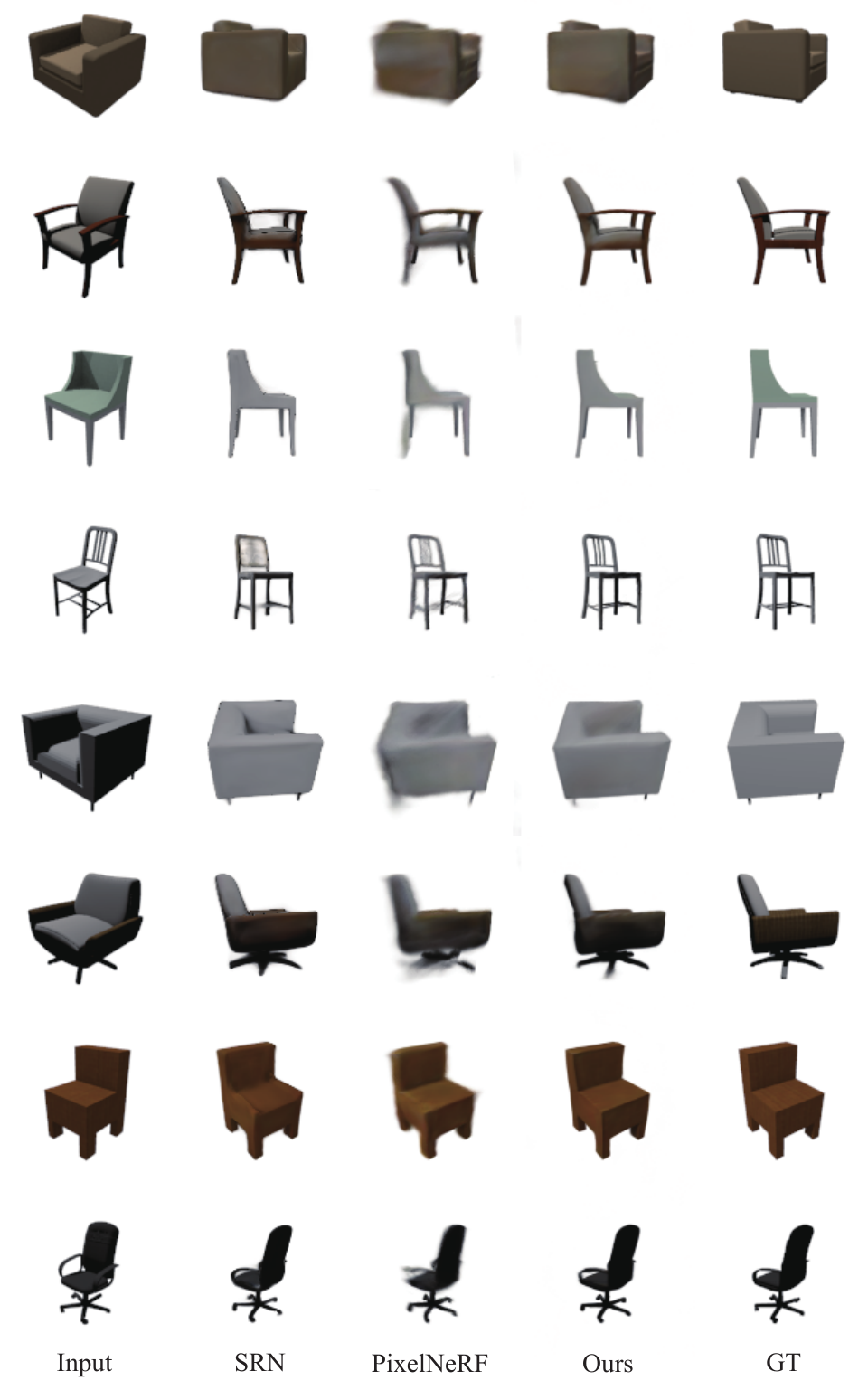}
\caption{\boldstart{Category-specific view synthesis results on chairs.}}
\label{fig:chairs}
\end{figure*}

\begin{figure*}[t]
\centering
\includegraphics[width=0.9\textwidth]{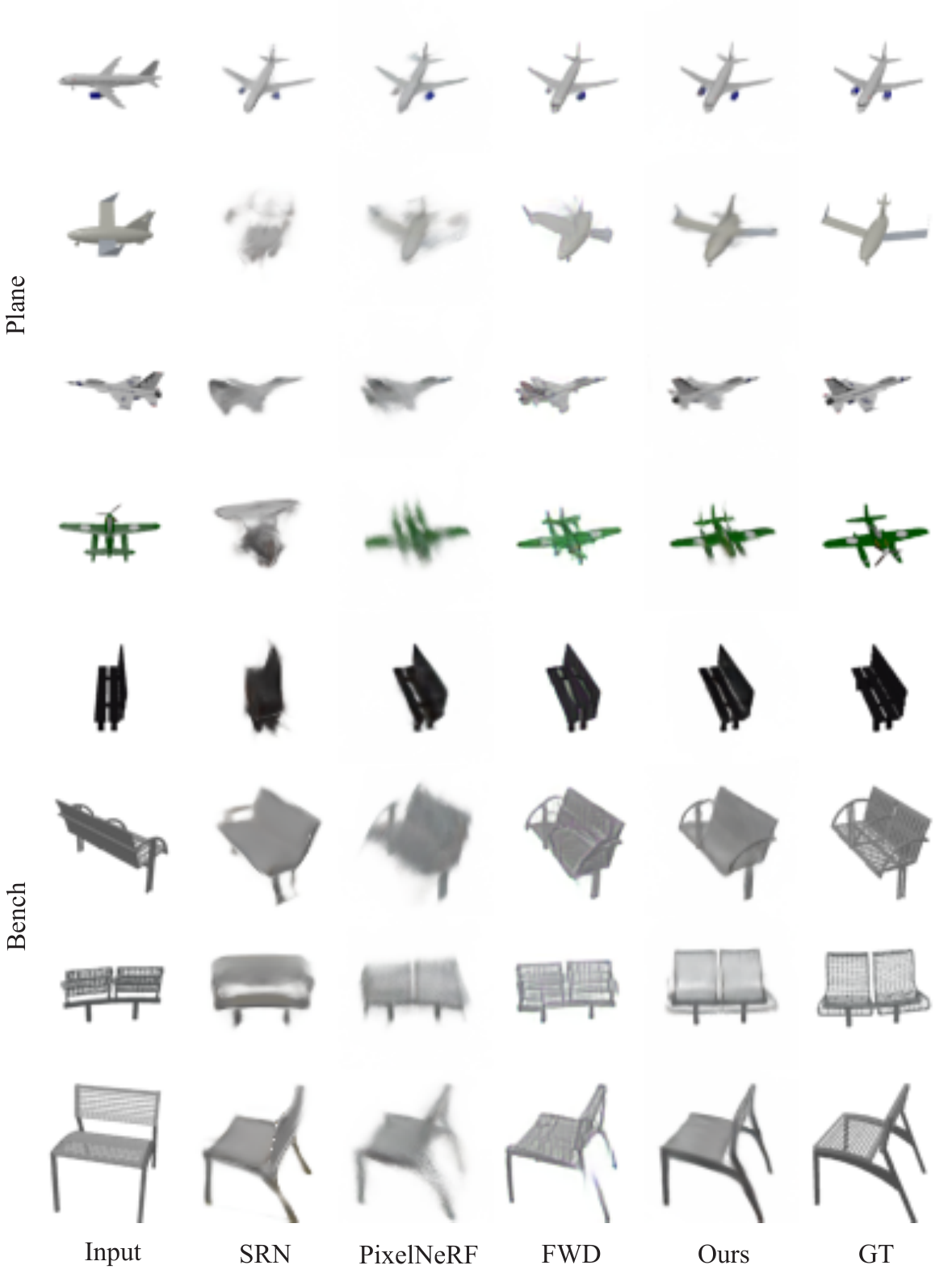}
\caption{\boldstart{Category-agnostic view synthesis results on the NMR dataset.} }
\label{fig:nmr_1}
\end{figure*}

\begin{figure*}[t]
\centering
\includegraphics[width=0.9\textwidth]{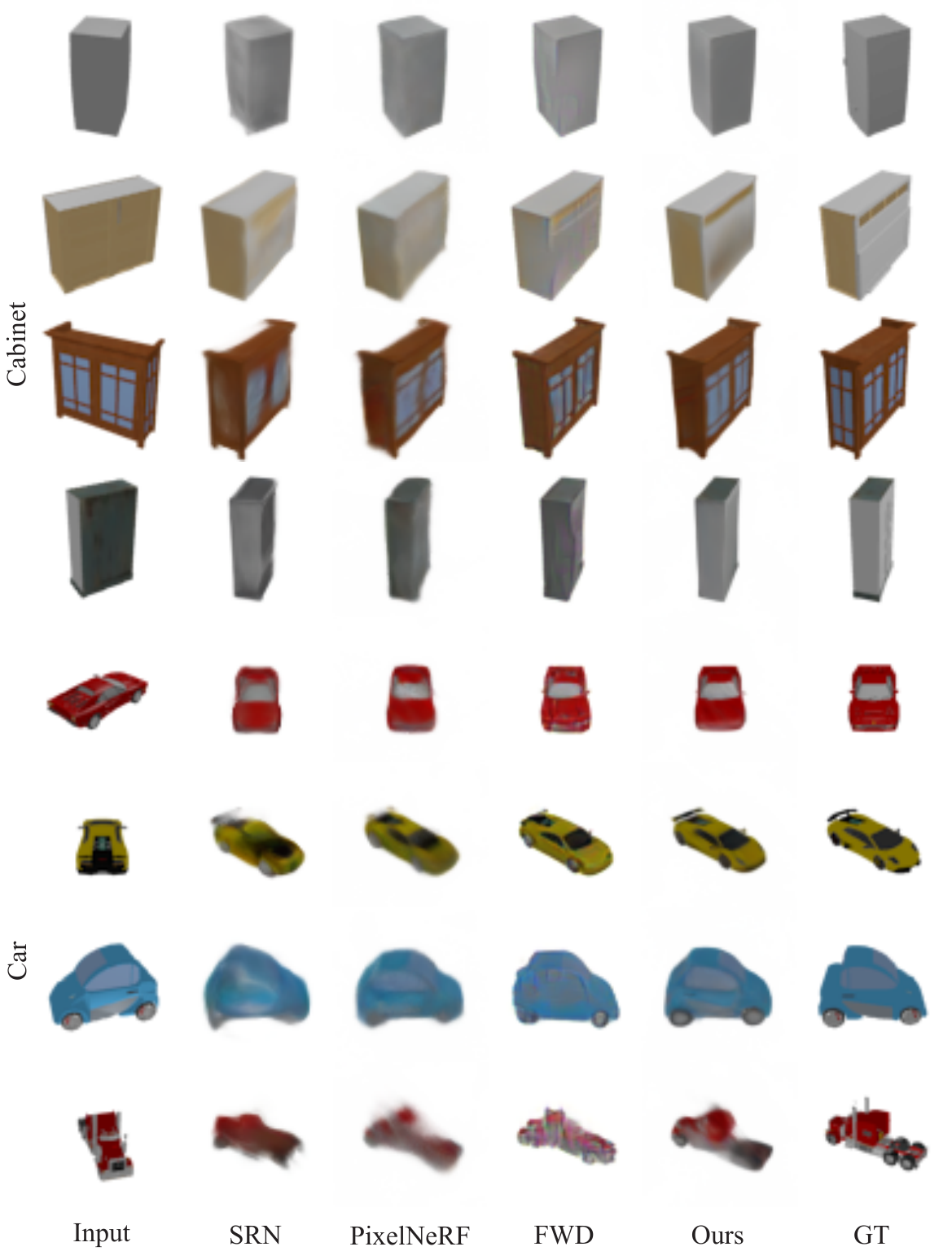}
\caption{\boldstart{Category-agnostic view synthesis results on the NMR dataset.} }
\label{fig:nmr_2}
\end{figure*}

\begin{figure*}[t]
\centering
\includegraphics[width=0.9\textwidth]{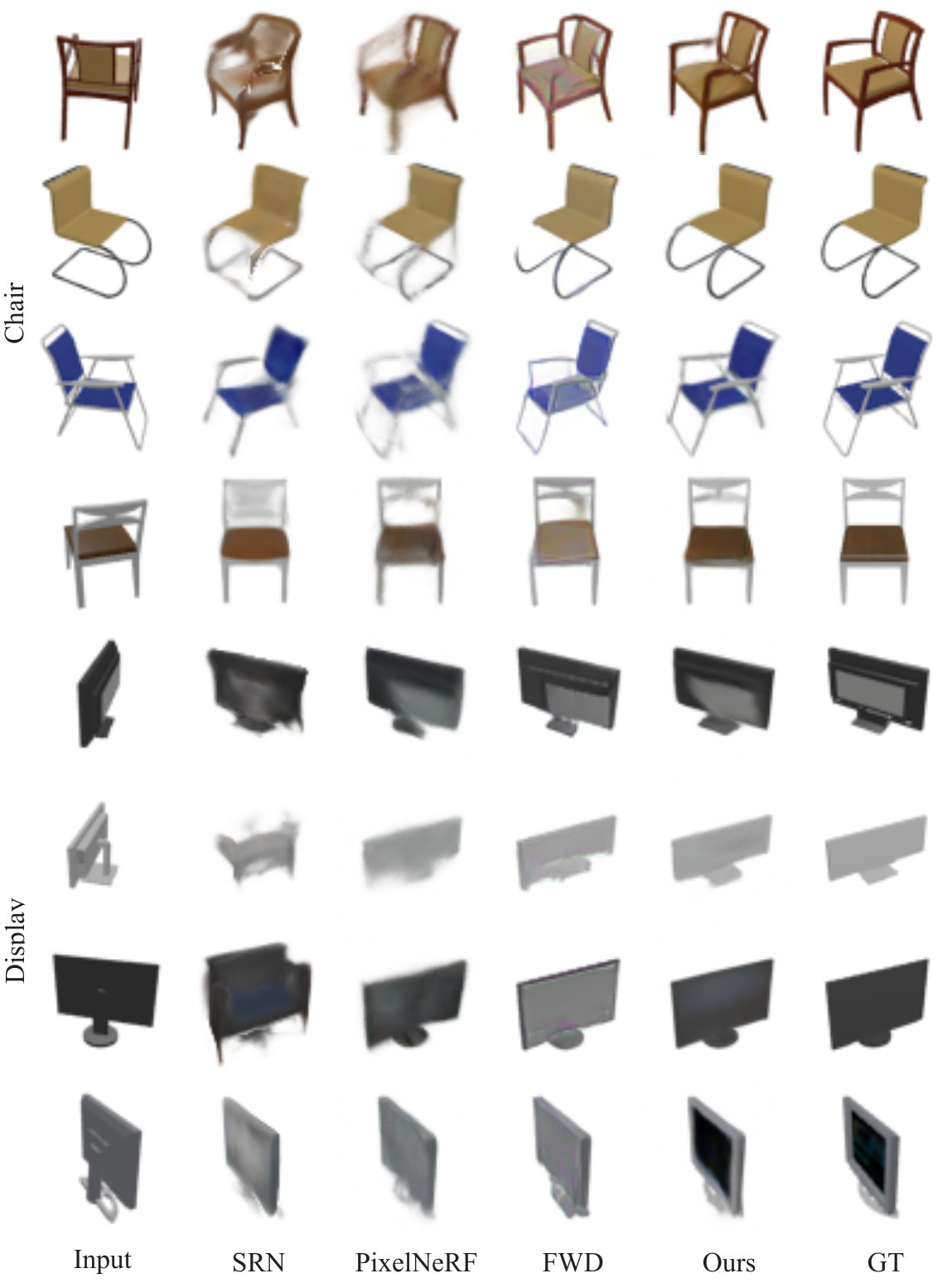}
\caption{\boldstart{Category-agnostic view synthesis results on the NMR dataset.} }
\label{fig:nmr_3}
\end{figure*}

\begin{figure*}[t]
\centering
\includegraphics[width=0.9\textwidth]{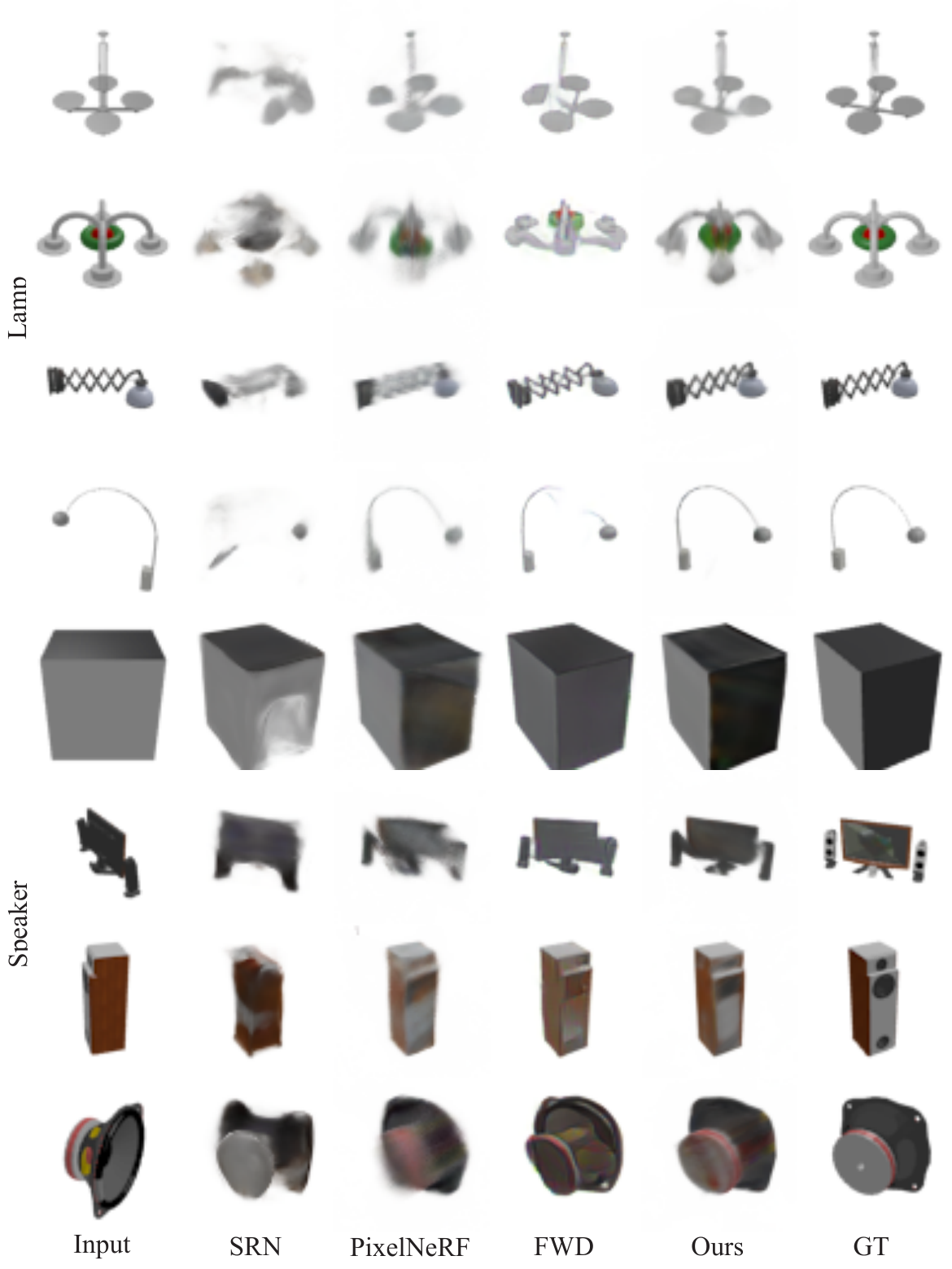}
\caption{\boldstart{Category-agnostic view synthesis results on the NMR dataset.} }
\label{fig:nmr_4}
\end{figure*}

\begin{figure*}[t]
\centering
\includegraphics[width=0.9\textwidth]{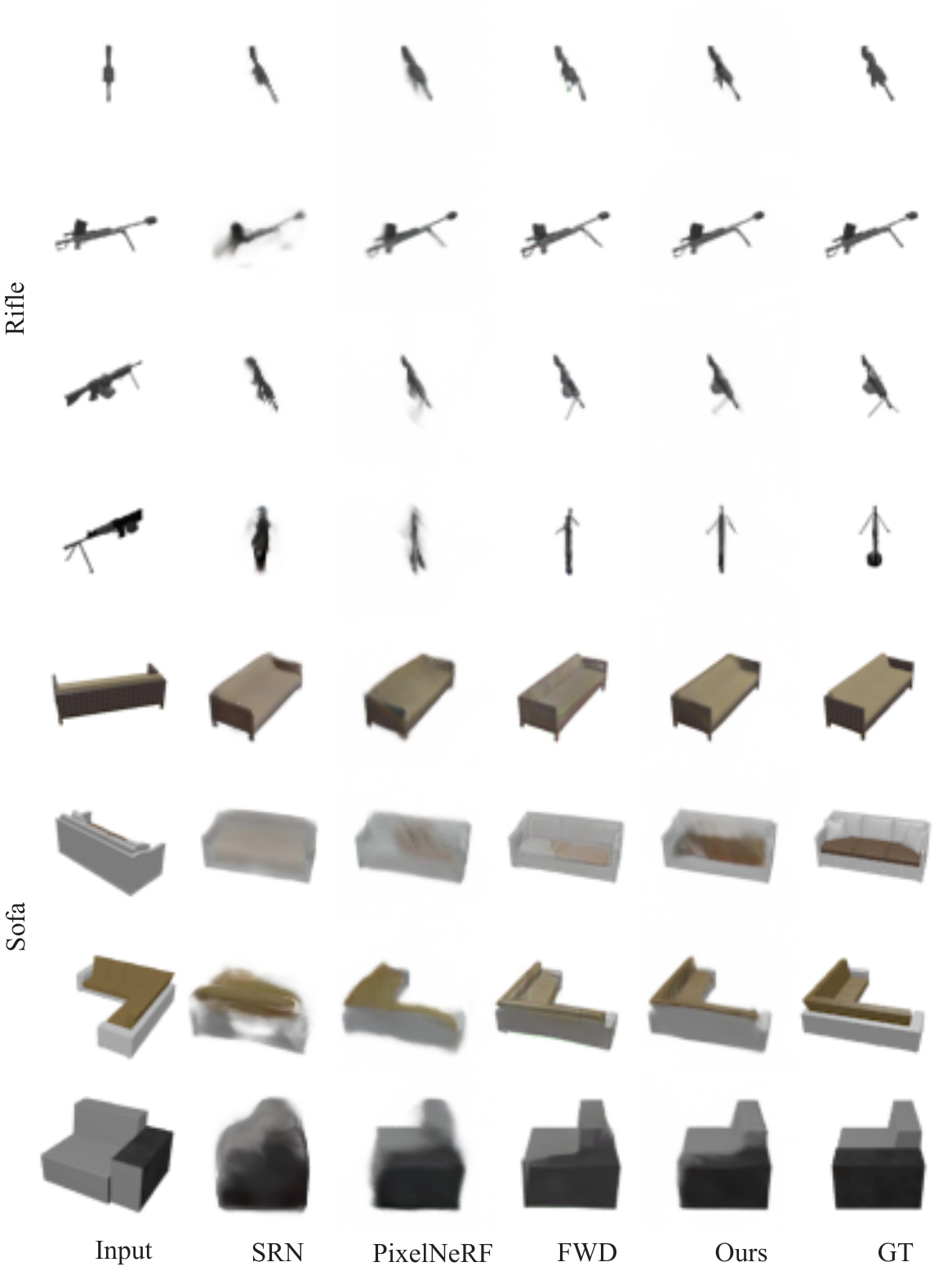}
\caption{\boldstart{Category-agnostic view synthesis results on the NMR dataset.} }
\label{fig:nmr_5}
\end{figure*}

\begin{figure*}[t]
\centering
\includegraphics[width=0.9\textwidth]{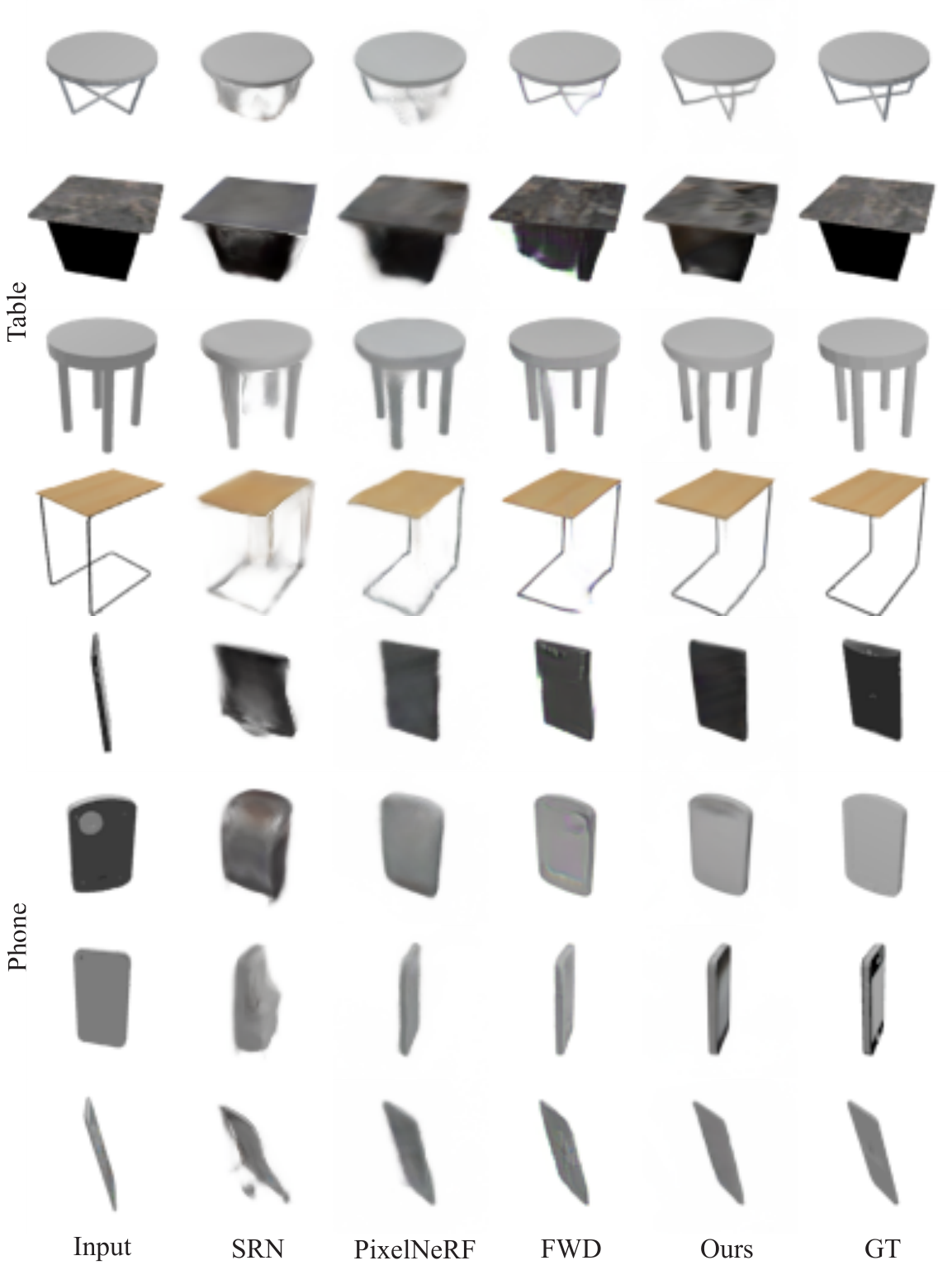}
\caption{\boldstart{Category-agnostic view synthesis results on the NMR dataset.} }
\label{fig:nmr_6}
\end{figure*}

\begin{figure*}[t]
\centering
\includegraphics[width=0.9\textwidth]{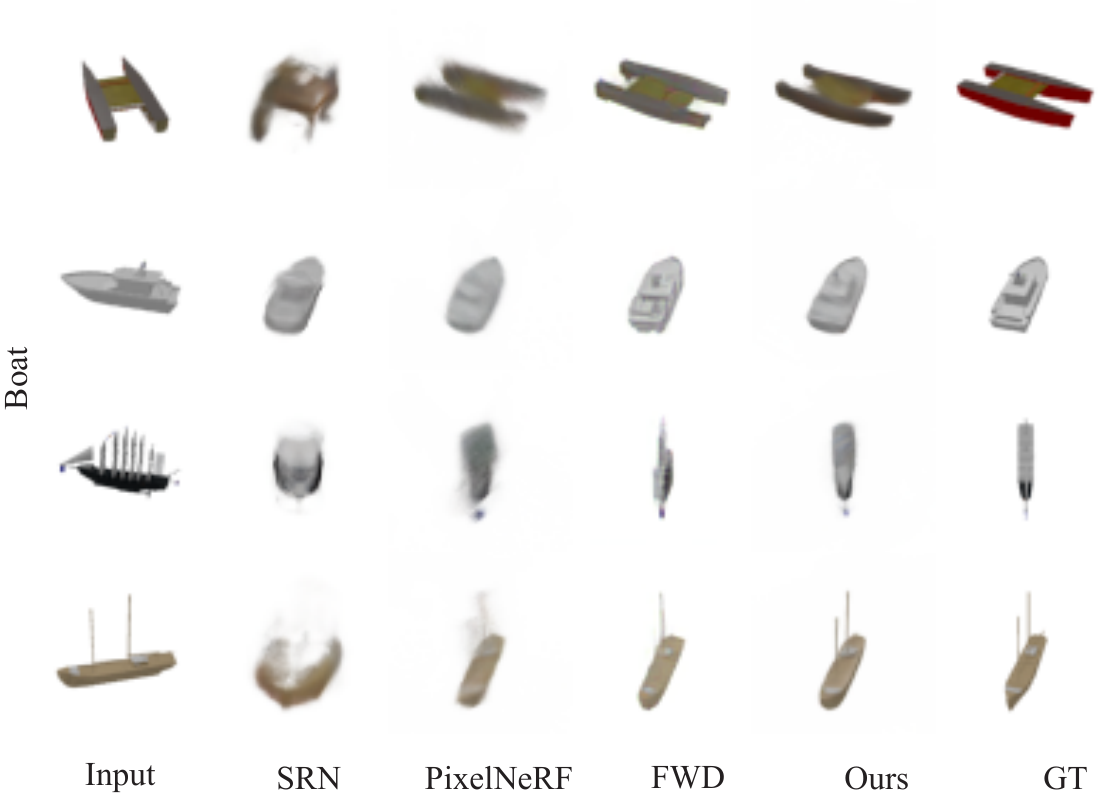}
\caption{\boldstart{Category-agnostic view synthesis results on the NMR dataset.} }
\label{fig:nmr_7}
\end{figure*}

We include extra results on category-specific view synthesis in Fig.~\ref{fig:cars} and~\ref{fig:chairs}, results on category-agnostic view synthesis in Fig.~\ref{fig:nmr_1}-\ref{fig:nmr_7}.
Note that we choose the target views where most pixels are not visible in the input view to better compare the rendering quality of each method on occluded regions.
The video results (in GIF format) are also provided on website (\url{https://cseweb.ucsd.edu/~viscomp/projects/VisionNeRF/supplementary.html}).



{\small
\bibliographystyle{ieee_fullname}
\bibliography{egbib}
}

\end{document}